%% file: iros2018_perception_aware_mpc.tex
\documentclass[letterpaper, 10pt, conference]{ieeeconf}  % Comment this line out if you need a4paper

\usepackage{balance}

\usepackage{subfig}
\usepackage{booktabs}
\usepackage{color}
\usepackage{footmisc}
\usepackage[bookmarks=true,hyperfootnotes=false]{hyperref}

\usepackage[binary-units=true]{siunitx}
\usepackage{graphicx}
\usepackage{bm}
\usepackage{amsmath} 
\usepackage{amssymb}
\usepackage{amsfonts}
\usepackage{mathtools}
\usepackage{footmisc}
\usepackage{todonotes}
\input{chapters/notation.tex}

\IEEEoverridecommandlockouts                              % This command is only needed if 
\newcommand{\rom}[1]{\uppercase\expandafter{\romannumeral #1\relax}}
\newcommand{\darkgrayed}[1]{\textcolor{darkgray}{#1}}
\makeatletter
\newcommand*\titleheader[1]{\gdef\@titleheader{#1}}
\AtBeginDocument{%
  \let\st@red@title\@title
  \def\@title{%
    \vskip-3em
    \bgroup\normalfont\large\centering\@titleheader\par\egroup
    \vskip1.5em\st@red@title}
}
\makeatother

\titleheader{\darkgrayed{This paper has been accepted for publication at the IEEE/RSJ International Conference on Intelligent Robots and Systems (IROS), Madrid, 2018.
\copyright IEEE}}

\title{\LARGE \bf
PAMPC: Perception-Aware Model Predictive Control for Quadrotors
}

\author{Davide Falanga$^*$, Philipp Foehn$^*$, Peng Lu, and Davide Scaramuzza 
\thanks{$^*$ These authors contributed equally to this manuscript.}
\thanks{This research was supported by the National Centre of Competence in Research (NCCR) Robotics, the SNSF-ERC Starting Grant, the DARPA FLA program. The authors are with the Robotics and Perception Group, Dep. of Informatics, University of Zurich, and Dep. of Neuroinformatics, University of Zurich and ETH Zurich, Switzerland---\url{http://rpg.ifi.uzh.ch}.}%
}

\begin{document}

\maketitle
\thispagestyle{empty}
\pagestyle{empty}

\input{chapters/abstract.tex}
\input{chapters/introduction.tex}
\input{chapters/problem_formulation.tex}
\input{chapters/methodology.tex}
\input{chapters/optimization.tex}
\input{chapters/experiments.tex}
\input{chapters/discussion.tex}
\input{chapters/conclusions.tex}

\bibliographystyle{IEEEtran}
\balance
\bibliography{references}

\end{document}

%% file: chapters/notation.tex
\newcommand{\bVec}[1]{\mathbf{#1}}
\newcommand{\quantityWithFrames}[4]{_{#1}#2_{#3#4}}

\newcommand{\CameraFrame}[0]{\ensuremath{C}}
\newcommand{\BodyFrame}[0]{\ensuremath{B}}
\newcommand{\WorldFrame}[0]{\ensuremath{W}}

\newcommand{\FocalLengthx}[0]{f_{x}}
\newcommand{\FocalLengthy}[0]{f_{y}}
\newcommand{\gravity}[0]{_W\bVec{g}}

\newcommand{\TWB}[0]{\quantityWithFrames{}{T}{\WorldFrame}{\BodyFrame}}

\newcommand{\bodyPosInWorld}[0]{\quantityWithFrames{}{\bVec{p}}{\WorldFrame}{\BodyFrame}}
\newcommand{\bodyPosInWorldWithReferenceFrame}[0]{\quantityWithFrames{\WorldFrame}{\bVec{p}}{\WorldFrame}{\BodyFrame}}
\newcommand{\bodyPosInWorldDot}[0]{\quantityWithFrames{}{\bVec{\dot{p}}}{\WorldFrame}{\BodyFrame}}

\newcommand{\bodyQuatInWorld}[0]{\quantityWithFrames{}{\bVec{q}}{\WorldFrame}{\BodyFrame}}
\newcommand{\bodyQuatInWorldDot}[0]{\quantityWithFrames{}{\bVec{\dot{q}}}{\WorldFrame}{\BodyFrame}}

\newcommand{\bodyLinVelInWorld}[0]{\quantityWithFrames{}{\bVec{v}}{\WorldFrame}{\BodyFrame}}
\newcommand{\bodyLinVelInWorldDot}[0]{\quantityWithFrames{}{\bVec{\dot{v}}}{\WorldFrame}{\BodyFrame}}

\newcommand{\bodyAngVelInBody}[0]{\quantityWithFrames{}{\bVec{\Omega}}{\BodyFrame}{}}
\newcommand{\bodyAngVelx}[0]{\omega_x}
\newcommand{\bodyAngVely}[0]{\omega_y}
\newcommand{\bodyAngVelz}[0]{\omega_z}

\newcommand{\bodyx}[0]{\bVec{x}_\BodyFrame}
\newcommand{\bodyy}[0]{\bVec{y}_\BodyFrame}
\newcommand{\bodyz}[0]{\bVec{z}_\BodyFrame}

\newcommand{\worldx}[0]{\bVec{x}_\WorldFrame}
\newcommand{\worldy}[0]{\bVec{y}_\WorldFrame}
\newcommand{\worldz}[0]{\bVec{z}_\WorldFrame}

\newcommand{\camerax}[0]{\bVec{x}_\CameraFrame}
\newcommand{\cameray}[0]{\bVec{y}_\CameraFrame}
\newcommand{\cameraz}[0]{\bVec{z}_\CameraFrame}
\newcommand{\TBC}[0]{\quantityWithFrames{}{T}{\BodyFrame}{\CameraFrame}}

\newcommand{\camPosInBody}[0]{\quantityWithFrames{}{\bVec{p}}{\BodyFrame}{\CameraFrame}}

\newcommand{\camQuatInBody}[0]{\quantityWithFrames{}{\bVec{q}}{\BodyFrame}{\CameraFrame}}
\newcommand{\camQuatInBodyInv}[0]{\quantityWithFrames{}{\bVec{q}^{-1}}{\BodyFrame}{\CameraFrame}}

\newcommand{\camLinVelInCamera}[0]{\quantityWithFrames{\CameraFrame}{\bVec{v}}{\WorldFrame}{\CameraFrame}}

\newcommand{\camAngVelInCamera}[0]{\quantityWithFrames{}{\bVec{\Omega}}{\CameraFrame}{}}

\newcommand{\imageCoordinates}[0]{\left(u,v\right)}

\newcommand{\featurePosInWorld}[0]{\quantityWithFrames{\WorldFrame}{\bVec{p}}{f}{}}

\newcommand{\featurePosInWorldx}[0]{\quantityWithFrames{\WorldFrame}{p}{fx}{}}
\newcommand{\featurePosInWorldy}[0]{\quantityWithFrames{\WorldFrame}{p}{fy}{}}
\newcommand{\featurePosInWorldz}[0]{\quantityWithFrames{\WorldFrame}{p}{fz}{}}

\newcommand{\featurePosInCamera}[0]{\quantityWithFrames{\CameraFrame}{\bVec{p}}{f}{}}
\newcommand{\featurePosInCamerax}[0]{\quantityWithFrames{\CameraFrame}{p}{fx}{}}
\newcommand{\featurePosInCameray}[0]{\quantityWithFrames{\CameraFrame}{p}{fy}{}}
\newcommand{\featurePosInCameraz}[0]{\quantityWithFrames{\CameraFrame}{p}{fz}{}}

\newcommand{\featurePosInCameraDot}[0]{\quantityWithFrames{\CameraFrame}{\dot{\bVec{p}}}{f}{}}
\newcommand{\featurePosInCameraxDot}[0]{\quantityWithFrames{\CameraFrame}{\dot{p}}{fx}{}}
\newcommand{\featurePosInCamerayDot}[0]{\quantityWithFrames{\CameraFrame}{\dot{p}}{fy}{}}
\newcommand{\featurePosInCamerazDot}[0]{\quantityWithFrames{\CameraFrame}{\dot{p}}{fz}{}}

\newcommand{\featureVecInImage}[0]{\bVec{s}}
\newcommand{\featureVecInImageDot}[0]{\bVec{\dot{s}}}

\newcommand{\skewQuaternion}[1]{\Lambda \left(#1\right)}

\newcommand{\StateVector}[0]{\bVec{x}}
\newcommand{\InputVector}[0]{\bVec{u}}

\newcommand{\collectiveThrust}[0]{c}
\newcommand{\singleRotorThrust}[1]{f_{#1}}
\newcommand{\collectiveThrustVector}[0]{\bVec{c}}
\newcommand{\mass}[0]{m}

\newcommand{\PerceptionState}[0]{\bVec{z}}
\newcommand{\PerceptionParameters}[0]{\bVec{\sigma}}

%% file: chapters/abstract.tex
\begin{abstract}
We present the first perception-aware model predictive control framework for quadrotors that unifies control and planning with respect to action and perception objectives.
Our framework leverages numerical optimization to compute trajectories that satisfy the system dynamics and require control inputs within the limits of the platform.
Simultaneously, it optimizes perception objectives for robust and reliable sensing by maximizing the visibility of a point of interest and minimizing its velocity in the image plane.
Considering both perception and action objectives for motion planning and control is challenging due to the possible conflicts arising from their respective requirements.
For example, for a quadrotor to track a reference trajectory, it needs to rotate to align its thrust with the direction of the desired acceleration.
However, the perception objective might require to minimize such rotation to maximize the visibility of a point of interest.
A model-based optimization framework, able to consider both perception and action objectives and couple them through the system dynamics, is therefore necessary.
Our perception-aware model predictive control framework works in a receding-horizon fashion by iteratively solving a non-linear optimization problem.
It is capable of running in real-time, fully onboard our lightweight, small-scale quadrotor using a low-power ARM computer, together with a visual-inertial odometry pipeline.
We validate our approach in experiments demonstrating (\rom{1}) the conflict between perception and action objectives, and (\rom{2}) improved behavior in extremely challenging lighting conditions.
\end{abstract}

%% file: chapters/introduction.tex
\section*{Supplementary material}
\noindent Video: \url{https://youtu.be/9vaj829vE18}

\noindent Code: \url{https://github.com/uzh-rpg/rpg_mpc}

\section{Introduction}
Thanks to the progresses in perception algorithms, the availability of low-cost cameras, and the increased computational power of small-scale computers, vision-based perception has recently emerged as the de facto standard in onboard sensing for micro aerial vehicles.
This made it possible to replicate some of the impressive quadrotor maneuvers seen in the last decade~\cite{Mellinger10iser,Mellinger11icra,Mueller15tro,Brescianini13iros}, which relied on motion-capture systems, using only onboard sensing, such as cameras and IMUs~\cite{Falanga17icra,Loianno17ral,Su17iser}.

Cameras have a number of advantages over other sensors in terms of weight, cost, size, power consumption and field of view.
However, vision-based perception has severe limitations: it can be intermittent and its accuracy is strongly affected by both the environment (e.g., texture distribution, light conditions) and motion of the robot (e.g., motion blur, camera pointing direction, distance from the scene). 
This means that one cannot always replace motion-capture systems with onboard vision, since the motion of a camera can negatively affect the quality of the estimation, posing hard bounds on the agility of the robot. 
On the other hand, perception can benefit from the robot motion if it is planned considering the necessities and the limitations of onboard vision. 
For example, to pass through a narrow gap while localizing with respect to it using an onboard camera, it is necessary to guarantee that the gap is visible at all times.
Similarly, to navigate through an unknown environment, it is necessary to guarantee that the camera always points towards texture-rich regions.

To fully leverage the agility of autonomous quadrotors, it is necessary to create synergy between perception and action by considering them jointly as a single problem.
\begin{figure}[t]
	\centering
	\includegraphics[width=\linewidth]{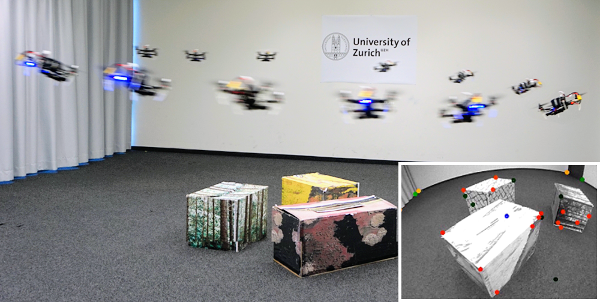}
	\caption{An example application of our PAMPC, where a quadrotor is asked to fly at \SI{3}{\meter \per \second} around a region of interest while keeping it visible in the field of view of its camera.}
	\label{fig:eye_catching}
\end{figure}

\subsection{Contributions}
Model Predictive Control (MPC) has become increasingly popular for quadrotor control~\cite{Kamel16Arxiv,Neunert16icra,Bangura14ifac} thanks to its capability of simultaneously dealing with different constraints and objectives through optimization.
In this work, we present an MPC algorithm for quadrotors able to optimize both action and perception objectives.

Our framework satisfies the robot dynamics and computes feasible trajectories with respect to the input saturations.
Such trajectories are not constrained to specific time or space parametrization (e.g., polynomials in time or splines), and tightly couple perception and action.
To do so, perception objectives aimed at rendering vision-based estimation more robust are taken into account in the optimization problem.
Such objectives are the visibility of a point of interest the robot needs to maintain in the image, and the minimization of the velocity of its projection onto the image plane.
The main challenge in this is to simultaneously cope with action (e.g., dynamics, underactuation, saturations) and perception objectives, due to the potential conflicts between them.

To solve this problem, we leverage numerical optimization to compute trajectories that are optimal with respect to a cost function considering both the dynamics of the robot and the quality of perception.
To fully exploit the agility of a quadrotor, we incorporate perception objectives into the optimization problem not as constraints, but rather as components to be optimized.
This results in a perception-aware framework which is intrinsically tailored to agile navigation, since the optimizer can trade off between perception and action objectives (cf. Fig.~\ref{fig:eye_catching}, depicting fast circle flight while adjusting the heading to look at a point of interest).
Furthermore, considering perception in the cost function reduces the computation load of the model predictive control pipeline, allowing it to run in real-time on a low-power onboard computer.
Our approach does not depend on the task and can potentially provide benefits to a large variety of applications, such as vision-based localization, target tracking, visual servoing, and obstacle detection.
We validate our perception-aware model predictive control framework in real-world experiments using a small-scale, lightweight quadrotor platform.

\subsection{Related Work}
The aforementioned shift from offboard to onboard sensing based on cameras resulted in an increased number of works trying to connect perception and action.

In~\cite{Penin17iros}, the authors proposed a method to compute minimum-time trajectories that take into account the limited field of view of a camera to guarantee visibility of points of interests.
Such a method requires the trajectory to be parametrized as a B-spline polynomial, constraining the kind of motion the robot can perform.
Also, perception is included in the planning problem as hard constraint, posing an upper-bound to the agility of the robot since such constraints must be satisfied at all times.
Furthermore, the velocity of the projection of the points of interest in the image is not taken into account.
Finally, the algorithm was not suited for real-time control of a quadrotor, and was only tested in simulations..

In~\cite{Spica17ijrr}, the authors focused on combining visual servoing with active Structure from Motion and proposed a solution to modify the trajectory of a camera in order to increase the quality of the reconstruction.
In such a work, a trajectory for the tracked features in the image plane was required, and the null space of the visual servoing task was exploited in order to render it possible for such feature to track the desired trajectory.
Furthermore, the authors did not consider the underactuation of the robot, which can significantly lower the performance of the overall task due to potentially conflicting dynamics and perception objectives.

In~\cite{Costante17ISRR} and~\cite{Forster14rss}, information gain was used to bridge the gap between perception and action.
In the first work, the authors tackled the problem of selecting trajectories that minimize the pose uncertainty by driving the robot toward regions rich of texture.
In the second work, a technique to minimize the uncertainty of a dense 3D reconstruction based on the scene appearance was proposed.
In both works, however, near-hover quadrotor flight was considered, and the underactation of the platform was not taken into account.

In~\cite{Sheckells16iros}, a hybrid visual servoing technique for differentially flat systems was presented.
A polynomial parameterization of the flat outputs of the system was required, and due to the computational load required by the designed optimization framework, an optimal trajectory was computed in advance and never replanned.
This did not allow coping with external disturbances and unmodelled dynamics, which during the execution of the trajectory can lead to behaviours different from the expected one.

In~\cite{Naegeli17ral} and~\cite{Naegeli17siggraph}, a real-time motion planning method for aerial videography was presented.
In these works, the main goal was to optimize the viewpoint of a pan-tilt camera carried by an aerial robot in order to improve the quality of the video recordings.
Both works were mainly targeted to cinematography, therefore they considered objectives such as the size of a target of interest and its visibility.
Conversely, we target robotic sensing and consider objectives aimed at facilitating vision-based perception.

In~\cite{Potena17ecmr}, the authors proposed a two-step approach for target-aware visual navigation.
First, position-based visual servoing was exploited to find a trajectory minimizing the reprojection error of a landmark of interest.
Then, a model predictive control pipeline was used to track such a trajectory.
Conversely, we solve the trajectory optimization and tracking within a single framework.
Additionally, that work only aimed at rendering the target visible, but did not take into account that, due to the motion of the camera, it might not be detectable because of motion blur.
We cope with this problem by considering in the optimization problem the velocity of the projection of the point of interest in the image plane.

\subsection{Structure of the Paper}
The remainder of this paper is organized as follows. 
In Sec.~\ref{sec:problem_formulation} we provide the general formulation of the problem.
In Sec.~\ref{sec:methodology} we derive the model for the dynamics of the projection of a 3D point into the image plane for the case of a quadrotor equipped with a camera.
In Sec.~\ref{sec:optimization} we present our perception-aware optimization framework, describing the objectives and the constraints it takes into account.
In Sec.~\ref{sec:experiments} we validate our approach in different real-world experiments showcasing the capability of our framework.
In Sec.~\ref{sec:discussion} we discuss our approach and provide additional insights and in Sec.~\ref{sec:conclusions} we draw the conclusions.

%% file: chapters/problem_formulation.tex
\section{Problem Formulation} \label{sec:problem_formulation}
For truly autonomous robot navigation, two components are essential: 
(\rom{1}) perception, both of the ego-motion and of the surrounding environment; 
(\rom{2}) action, meant as the combination of motion planning and control algorithms.
A very wide literature is available for both of them.
However, they are rarely considered as a joint problem. 

The need for coupled perception and action can be easily explained.
To guarantee safety, accurate and robust perception is necessary.
Nevertheless, the quality of vision-based perception is strongly affected by the motion of the camera.
On the one hand, it can degrade its performance by not making it possible to extract sufficiently accurate information from images.
For example, lack of texture or blur due to camera motion can lead to algorithm failure.
On the other, the quality of vision-based perception can improve significantly if its limitations and requirements are considered, e.g. by rendering highly-textured areas visible in the image and by reducing motion blur.
Therefore it is necessary to create synergy between perception and action.

Let $\StateVector$ and $\InputVector$ be the state and input vectors of a robot, respectively.
Assume its dynamics to be described by a set of differential equations ${\dot{\StateVector} = \bm{f}\left(\StateVector,\InputVector\right)}$.
Furthermore, let $\PerceptionState$ be the state vector of the perception system (e.g., 3D points' projection onto the image plane), and $\PerceptionParameters$ a vector of parameters characterizing it (e.g., the focal length of the camera or its field of view).
The perception state and the robot state are coupled through the robot dynamics, namely ${\PerceptionState = \bm{f}_p\left(\StateVector, \InputVector, \PerceptionParameters\right)}$.
Given certain action objectives, we can define an action cost $\mathcal{L}_a\left(\StateVector, \InputVector\right)$.
Similarly, we can define a cost $\mathcal{L}_p\left(\PerceptionState\right)$ for the perception objectives.

We can then formulate the coupling of perception and action as an optimization problem:
\begin{align}
\begin{split}
\min_{\bVec{u}} \quad \int_{t_0}^{t_f} \mathcal{L}_a\left(\StateVector, \InputVector\right) + \mathcal{L}_p\left(\PerceptionState\right) dt& \\
\text{subject to} \qquad 
\bVec{r}(\StateVector, \InputVector,\PerceptionState) &= 0 \\
\bVec{h}(\StateVector, \InputVector,\PerceptionState) &\leq 0,
\label{eq:problem_formulation}
\end{split}
\end{align}
where $\bVec{r}(\StateVector, \InputVector,\PerceptionState)$ and $\bVec{h}(\StateVector, \InputVector,\PerceptionState)$ represent equality and inequality constraints that the solution should satisfy for perception, action, or both of them simultaneously.

%% file: chapters/methodology.tex
\section{Methodology} \label{sec:methodology}
Any computer vision algorithm aimed at providing a robot with the information necessary for navigation (e.g., pose estimation, obstacle detection, etc) has two fundamental requirements.
First, the points of interest used by the algorithm to provide the aforementioned information must be visible in the image.
For example, such points can be the landmarks used for pose estimation by visual odometry algorithms, or the points belonging to an object for obstacle detection.
If such points are not visible while the robot is moving, there is no way the algorithm can cope with the absence of information.
Second, such points of interest must be clearly recognizable in the image.
Depending on the motion of the camera and the distance from the scene, the projection of a 3D point onto an image can suffer from motion blur, making it very complicated, if not impossible, to extract meaningful information.
Therefore, the motion of the camera should be thoroughly planned to guarantee robust visual perception. 

Based on the considerations above, in this work we consider two perception objectives in our framework:~(\rom{1})~visibility of points of interest, and (\rom{2}) minimization of the velocity of their projection onto the image plane.
In the following, we study the relation between the motion of a quadrotor equipped with an onboard camera and the projection onto the image plane of a point in space.
Without loss of generality, we consider the case of a single 3D point of interest.
Our goal is to couple perception and action into an optimization framework by expressing the dynamics of its projection onto the image plane as a function of the state and input vectors of a quadrotor.

\subsection{Nomenclature}
\begin{figure}[t]
\centering
\includegraphics[width=\linewidth]{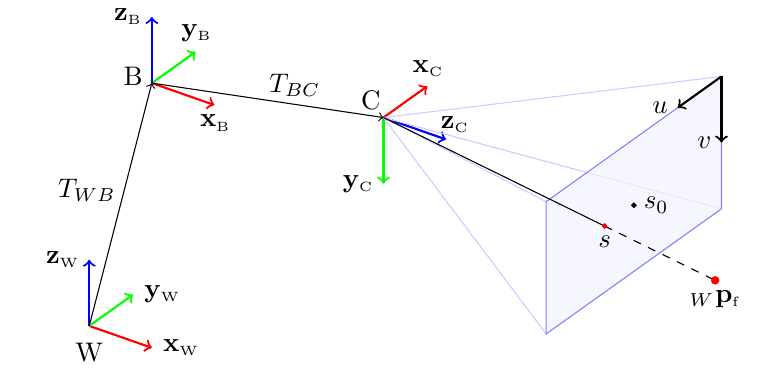}
\caption{A schematics representing the world frame $\WorldFrame$, the body frame $\BodyFrame$ and the camera frame $\CameraFrame$.
The position and orientation of $\BodyFrame$ with respect to $\WorldFrame$ is provided by $\TWB$.
The constant rigid body transformation $\TBC$ provides the extrinsics of the camera.
A feature located at $\featurePosInWorld$ is projected into the image plane onto a point of coordinates $\featureVecInImage$.
$\featureVecInImage_0$ represents the principal point.}
\label{fig:frames_schematics}
\end{figure}
In this work, we make use of a world frame $\WorldFrame$ with orthonormal basis ${\left\{ \worldx, \worldy, \worldz \right\}}$.
The quadrotor frame $\BodyFrame$, also referred to as the body frame, has orthonormal basis ${\left\{ \bodyx, \bodyy, \bodyz \right\}}$.
Finally, we assume the robot to be equipped with a camera, whose reference frame $\CameraFrame$ has orthonormal basis  ${\left\{ \camerax, \cameray, \cameraz \right\}}$.
Fig.~\ref{fig:frames_schematics} provides a clear overview about the reference frames.

Throughout this manuscript, we represent vectors as bold quantities having a prefix, representing the frame in which they are expressed, and a suffix, indicating the origin and the end of such a vector.
For example, the quantity $\bodyPosInWorldWithReferenceFrame$ represents the position of the body frame $\BodyFrame$ with respect to the world frame $\WorldFrame$, expressed in the world frame.
To simplify the notation, if a vector has no prefix, we assume it to be expressed in the first frame reported in the suffix (i.e., the frame where the vectors origin is).

We use quaternions to represent the orientation of a rigid body.
The time derivative of a quaternion $\bVec{q} = \left(q_w, q_x, q_y, q_z\right)^\intercal$ is given by ${\bVec{\dot{q}} = \frac{1}{2}\skewQuaternion{\bVec{\omega}} \cdot \bVec{q}}$, where the skew-symmetric matrix $\skewQuaternion{\bVec{\omega}}$ of a vector ${\bVec{\omega} = \left(\omega_x, \omega_y, \omega_z\right)}^\intercal$ is defined as:
\begin{equation}
	\skewQuaternion{\omega} = 
	\begin{bmatrix} 0 & -\bodyAngVelx & -\bodyAngVely & -\bodyAngVelz \\ \bodyAngVelx & 0 & \bodyAngVelz & -\bodyAngVely \\ \bodyAngVely & -\bodyAngVelz & 0 & \bodyAngVelx \\ \bodyAngVelz & \bodyAngVely & -\bodyAngVelx & 0 \end{bmatrix}.
\end{equation}
Finally, we use the operator $\odot$ to denote the multiplication between a quaternion and a vector.
More specifically, multiplying a vector $\bVec{v}$ with the quaternion $\bVec{q}$ means rotating $\bVec{v}$ by the rotation induced by $\bVec{q}$.
By doing so, we obtain a vector ${\bVec{v}^\prime = \bVec{v} \odot \bVec{q} = Q\bVec{v}}$ where:
\begin{equation}
Q = 
\begin{bmatrix} 
1 - 2q_y^2 - 2q_z^2    &    2(q_x q_y + q_w q_z)     &     2(q_x q_z - q_w q_y) \\
2(q_x q_y - q_w q_z)   &    1 - 2q_x^2 - 2q_z^2      &     2(q_y q_z + q_w q_x) \\
2(q_x q_z + q_w q_y)   &    2(q_y q_z - q_w q_x)     &     1 - 2q_x^2 - 2q_y^2
\end{bmatrix} \nonumber.
\end{equation}

\subsection{Quadrotor Dynamics}
Let ${\bodyPosInWorld = \left(p_x,p_y,p_z\right)^\intercal}$ and ${\bodyQuatInWorld = \left(q_w, q_x, q_y, q_z\right)^\intercal}$ be the position and the orientation of the body frame with respect to the world frame $\WorldFrame$, expressed in world frame, respectively (cf. Fig.~\ref{fig:frames_schematics}).
Additionally, let ${\bodyLinVelInWorld = \left(v_x, v_y, v_z\right)^\intercal}$ be the linear velocity of the body, expressed in world frame, and ${\bodyAngVelInBody = \left(\bodyAngVelx, \bodyAngVely, \bodyAngVelz\right)^\intercal}$ its angular velocity, expressed in the body frame.
Finally, let ${\collectiveThrustVector = \left(0, 0, \collectiveThrust\right)^\intercal}$ be the mass-normalized thrust vector, where ${\collectiveThrust = \left( \singleRotorThrust{1} + \singleRotorThrust{2} + \singleRotorThrust{3} + \singleRotorThrust{4}\right) / \mass}$, $\singleRotorThrust{i}$ is the thrust produced by the i-th motor, and $\mass$ is the mass of the vehicle.
In this work, we use the dynamical model of a quadrotor proposed in~\cite{Mueller15tro}:
\begin{align}
\begin{split}
\bodyPosInWorldDot &= \bodyLinVelInWorld \\
\bodyLinVelInWorldDot &= \gravity  + \bodyQuatInWorld \odot \collectiveThrustVector \\
\bodyQuatInWorldDot &= \frac{1}{2}\skewQuaternion{\bodyAngVelInBody} \cdot \bodyQuatInWorld
\label{eq:system_dynamics}
\end{split}
\end{align}
where $\gravity = \left(0, 0, -g\right)^\intercal$ is the gravity vector, with $g = \SI{9.81}{\meter \per \second \squared}$.
The state and the input vectors of the system are ${\StateVector = \left[\bodyPosInWorld, \bodyLinVelInWorld, \bodyQuatInWorld\right]^\intercal}$ and ${\InputVector = \left[\collectiveThrust, \bodyAngVelInBody^\intercal \right]^\intercal}$, respectively.

\subsection{Perception Objectives} \label{sec:perception_objectives}
Let $\featurePosInWorld = \left (\featurePosInWorldx, \featurePosInWorldy, \featurePosInWorldz \right)$ be the 3D position of a point of interest (landmark) in the world frame $\WorldFrame$ (cf. Fig.~\ref{fig:frames_schematics}).
We assume the body to be equipped with a camera having extrinsic parameters described by a constant rigid body transformation ${\TBC = \left[\camPosInBody,\camQuatInBody \right]}$, where $\camPosInBody$ and $\camQuatInBody$ are the position and the orientation of $\CameraFrame$ with respect to $\BodyFrame$.
The coordinates ${\featurePosInCamera= \left(\featurePosInCamerax, \featurePosInCameray, \featurePosInCameraz\right)^\intercal}$ of $\featurePosInWorld$ in the camera frame $\CameraFrame$ are given by:
\begin{equation} 
\begin{split} \label{eq:feature_position_in_camera_frame}
\featurePosInCamera= &\left (\bodyQuatInWorld~\camQuatInBody \right )^{-1} \odot \\& \left (\featurePosInWorld - \left (\bodyQuatInWorld \odot ~\camPosInBody + \bodyPosInWorld \right) \right ).
\end{split}
\end{equation} 
The point $\featurePosInCamera$ in camera frame is projected into the image plane coordinates $\featureVecInImage = \imageCoordinates^\intercal$ according to classical pinhole camera model~\cite{Szeliski10book}:
\begin{equation} \label{eq:camera_projection_model}
  u = \FocalLengthx \frac{\featurePosInCamerax}{\featurePosInCameraz}, \quad
  v = \FocalLengthy \frac{\featurePosInCameray}{\featurePosInCameraz}
\end{equation}
where $\FocalLengthx, \FocalLengthy$ are the focal lengths for pixel rows and columns, respectively.

To guarantee robust vision-based perception, the projection $\featureVecInImage$ of a point of interest $\featurePosInWorld$ should be as close as possible to the center of the image for two reasons.
First, keeping its projection in the center of the image results in the highest safety margins against external disturbances.
The second reason comes from the fact that the periphery of the image is typically characterized by a non-negligible distortion, especially for large field of view cameras.
A number of models for such distortion are available in the literature, as well as techniques to estimate their parameters to compensate the effects of the distortion.
However, such a compensation is never perfect and this can degrade the accuracy of the estimates.

As previously mentioned, in addition to rendering the point of interest visible in the image, we are interested in reducing the velocity of its projection onto the image plane.
We assume the point of interest to be static, but similar considerations apply to the case where such a point of interest moves with respect to the world frame.
To express the projection velocity as a function of the quadrotor state and input vectors, we can differentiate~\eqref{eq:camera_projection_model} with respect to time:
\begin{equation} \label{eq:derivative_of_camera_projection_model}
\begin{aligned} 
\dot{u} &= \FocalLengthx \frac{\featurePosInCameraxDot~\featurePosInCameraz -~ \featurePosInCamerax~\featurePosInCamerazDot}{\featurePosInCameraz^2}, \\
\dot{v} &= \FocalLengthy \frac{\featurePosInCamerayDot~\featurePosInCameraz -~ \featurePosInCameray~\featurePosInCamerazDot}{\featurePosInCameraz^2}.
\end{aligned}
\end{equation}
Eq.~\eqref{eq:derivative_of_camera_projection_model} can be written in a compact form as:
\begin{equation}\label{eq:derivative_of_pixel_coordinates}
\featureVecInImageDot = \begin{bmatrix}\dot{u} \\ \dot{v} \\ 0\end{bmatrix} = \begin{bmatrix} 0 & -\frac{\FocalLengthx}{\featurePosInCameraz^2} & 0 \\
													   \frac{\FocalLengthy}{\featurePosInCameraz^2} & 0 & 0 \\
													    0 & 0 & 0 \end{bmatrix} \left(\featurePosInCamera \times \featurePosInCameraDot \right).
\end{equation}
To compute the term $\featurePosInCameraDot $, we can differentiate~\eqref{eq:feature_position_in_camera_frame} with respect to time:
\begin{equation} 
\begin{split} \label{eq:derivative_of_feature_position_in_camera_frame}
\featurePosInCameraDot &= -\frac{1}{2}\skewQuaternion{\camAngVelInCamera}~\featurePosInCamera - \camLinVelInCamera,
\end{split}
\end{equation} 
where:
\begin{equation} \label{eq:camera_velocities_in_camera_frame}
\begin{aligned}
\camLinVelInCamera = &\left(\bodyQuatInWorld~\camQuatInBody\right)^{-1} \odot \\ &\left(\frac{1}{2}\skewQuaternion{\bodyAngVelInBody} \bodyQuatInWorld \odot \camPosInBody +\bodyLinVelInWorld \right), \\
\camAngVelInCamera &= \camQuatInBodyInv \odot \bodyAngVelInBody.
\end{aligned}
\end{equation}

\subsection{Action Objectives} \label{sec:action_objectives}
For a quadrotor to execute a desired task (e.g., reach a target position in space), a suitable trajectory has to be planned.
In this regard, for a quadrotor two objectives should be considered.

The first comes from the bounded inputs available to the system.
The thrust each motor can produce has both an upper and a lower bound, leading to a limited input vector $\InputVector$.
Therefore, denoting the subset of the allowed inputs as $\mathcal{U}$, the planned trajectory should be such that the condition $\InputVector(t) \in \mathcal{U}~\forall t$ can be satisfied.

The second objective to be considered comes from the underactuated nature of a quadrotor.
In the most common configuration, all the rotors point in the same direction, typically along the axis $\bodyz$ of the body.
This means that the robot can accelerate only in this direction.
Therefore, to move in the 3D space, it is necessary to exploit the system dynamics~\eqref{eq:system_dynamics} by coupling the translational and the rotational motions of the robot to follow the desired trajectory.

\subsection{Challenges} \label{sec:challenges}
The perception (Sec.~\ref{sec:perception_objectives}) and the action (Sec.~\ref{sec:action_objectives}) objectives previously described are both necessary for vision-based quadrotor navigation.
Considering them simultaneously is challenging due to the possible conflict among them.
Indeed, for a quadrotor to track a reference trajectory, it needs to rotate to align its thrust with the direction of the desired acceleration.
However, the perception objective might require to minimize such rotation to maximize the visibility of a point of interest.
A model-based optimization framework able to consider both perception and action objectives and couple them through the system dynamics is therefore necessary.

%% file: chapters/optimization.tex
\section{Model Predictive Control} \label{sec:optimization}

Formulating coupled perception and action as an optimization problem has the advantages of being able to satisfy the underactuated system dynamics and actuator constraints (i.e., input boundaries) and to minimize the predicted costs along a time horizon.
In contrast, classical control schemes are incapable of predicting costs and the corresponding trajectory (e.g., PID controllers) and guaranteeing input boundaries (PID, LQR).

The basic formulation of such an optimization is given in \eqref{eq:problem_formulation}, which in our case results in a non-linear program with quadratic costs.
This can then be approximated by a sequential quadratic program (SQP) where the solution of the non-linear program is iteratively approximated and used as a model predictive control (MPC).
To this regard, for the MPC to be effective, the optimization scheme has to run in real-time, at the desired control frequency.
To achieve this, we first discretize the system dynamics with a time step $dt$ for a time horizon $t_h$ into ${\bVec{x}_i~\forall i \in \left[1, N\right]}$ and ${\bVec{u}_i~\forall i \in \left[1, N-1\right]}$.
We define the time-varying state cost matrix as ${\mathcal{Q}_{x,i}~\forall i \in \left[1, N\right]}$.
Furthermore, the time-varying perception and input cost matrices are defined as $\mathcal{Q}_{p,i}$ and $\mathcal{R}_i$, ${\forall i \in \left[1, N-1\right]}$, respectively.
Finally, let $\PerceptionState = \left[ \featureVecInImage, \featureVecInImageDot \right]$ be the perception function.
It is important to recall that $\PerceptionState$ is a function of the quadrotor's state and input variables, as remarked in Eq.~\eqref{eq:feature_position_in_camera_frame} to~\eqref{eq:camera_velocities_in_camera_frame}.
The resulting cost function we consider is:
\begin{equation}
\begin{split}
\mathcal{L} = ~&\bar{\StateVector}_N^\intercal \mathcal{Q}_{x,N} \bar{\StateVector}_N ~+ \\ & + \sum_{i=1}^{N-1} \left(\bar{\StateVector}_i^\intercal \mathcal{Q}_{x,i} \bar{\StateVector}_i + \bar{\PerceptionState }_{i}^\intercal \mathcal{Q}_{p,i} \bar{\PerceptionState }_{i} + \bar{\InputVector}_i^\intercal \mathcal{R}_i \bar{\InputVector}_i\right),
\end{split}
\end{equation}
where the values $\bar{\StateVector}, \bar{\PerceptionState}, \bar{\InputVector}$ refer to the difference with respect to the reference of each value.
In our case, the reference value for $\PerceptionState$ is the null vector (i.e., center of the image and zero velocity) and the reference for the states and inputs are given by a target pose or a precomputed trajectory (that neglects the perception objectives).

The inputs $\InputVector$, consisting of $\collectiveThrust$ and $\bodyAngVelInBody$, as well as the velocity $\bodyLinVelInWorld$ are limited by the constraints:
\begin{gather}
\collectiveThrust_{min} \leq \collectiveThrust \leq \collectiveThrust_{max},  \\
-\Omega_{max} \leq \bodyAngVelInBody \leq \Omega_{max}, \\
-v_{max} \leq \bodyLinVelInWorld \leq v_{max},
\end{gather}
where $\collectiveThrust_{min}, \collectiveThrust_{max}, \Omega_{max}, v_{max} \in \mathcal{R}_+$.

To include the dynamics as in \eqref{eq:system_dynamics}, we use multiple shooting as transcription method and a Runge-Kutta integration scheme.
We refer the reader to \cite{Houska2011b} and \cite{Houska2011a} for more details on the transcription of the dynamics for optimization.

We approximate the solution of the optimization problem by executing one iteration at each control loop and use as initial state the most recent available estimate $\StateVector_{est}$ provided by a Visual-Inertial Odometry pipeline running onboard the vehicle (see Sec.~\ref{sec:experimental_setup}).
To achieve good approximations, it is important to run these iterations significantly faster than the discretization time of the problem and to keep the previous solution as initialization trajectory of the next optimization.
Such a SQP scheme leads to a fast convergence towards the exact solution, since the system is always close to the last linearization, and the deviation of each state $\StateVector_i$ between two iterations is very small.

%% file: chapters/experiments.tex
\section{Experiments} \label{sec:experiments}
In order to show the potential of our perception-aware model predictive control, we ran our approach onboard a small, vision-based, autonomous quadrotor.
We refer the reader to the attached video showcasing the experiments.
\begin{figure}[t]
	\centering
	\includegraphics[width=0.9\linewidth,trim={0 50 0 40},clip]{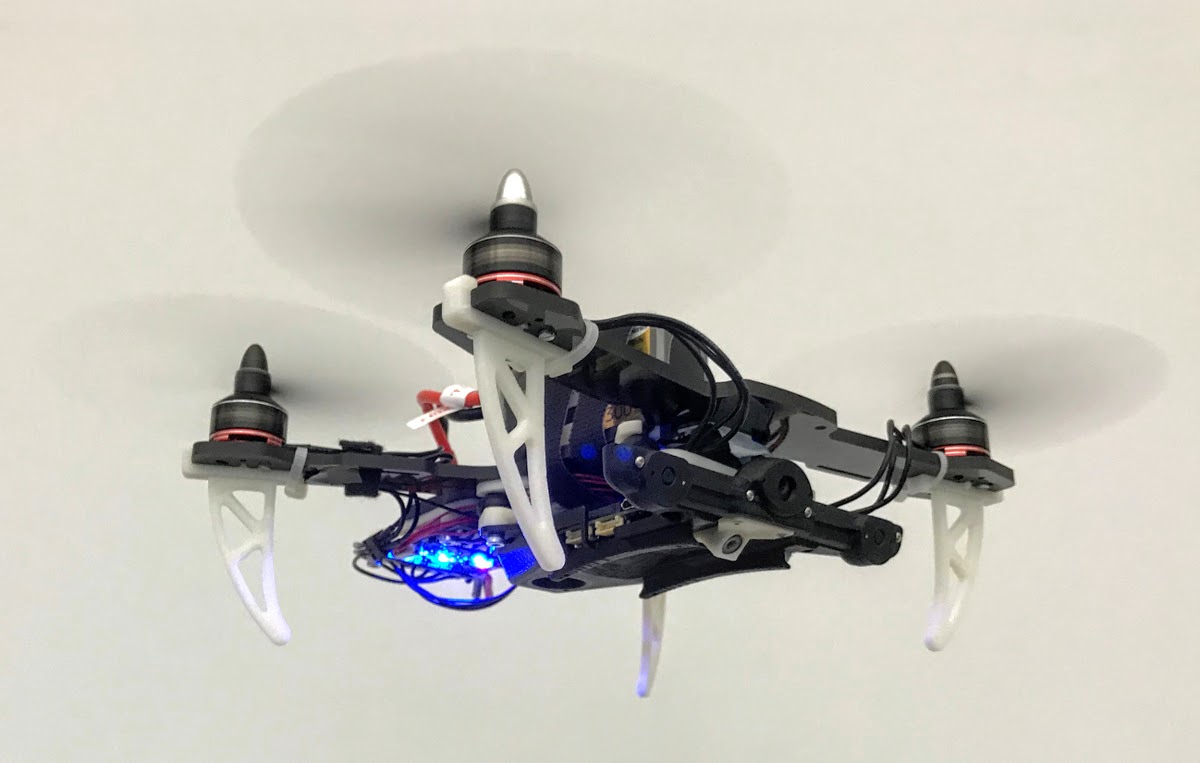}
	\caption{The quadrotor used for the experiments.}
	\label{fig:quadrotor_closeup}
\end{figure}
\begin{figure*}[t!]
	\centering
	\includegraphics[width=.21\textwidth,trim={5cm 1cm 7cm 7cm},clip]{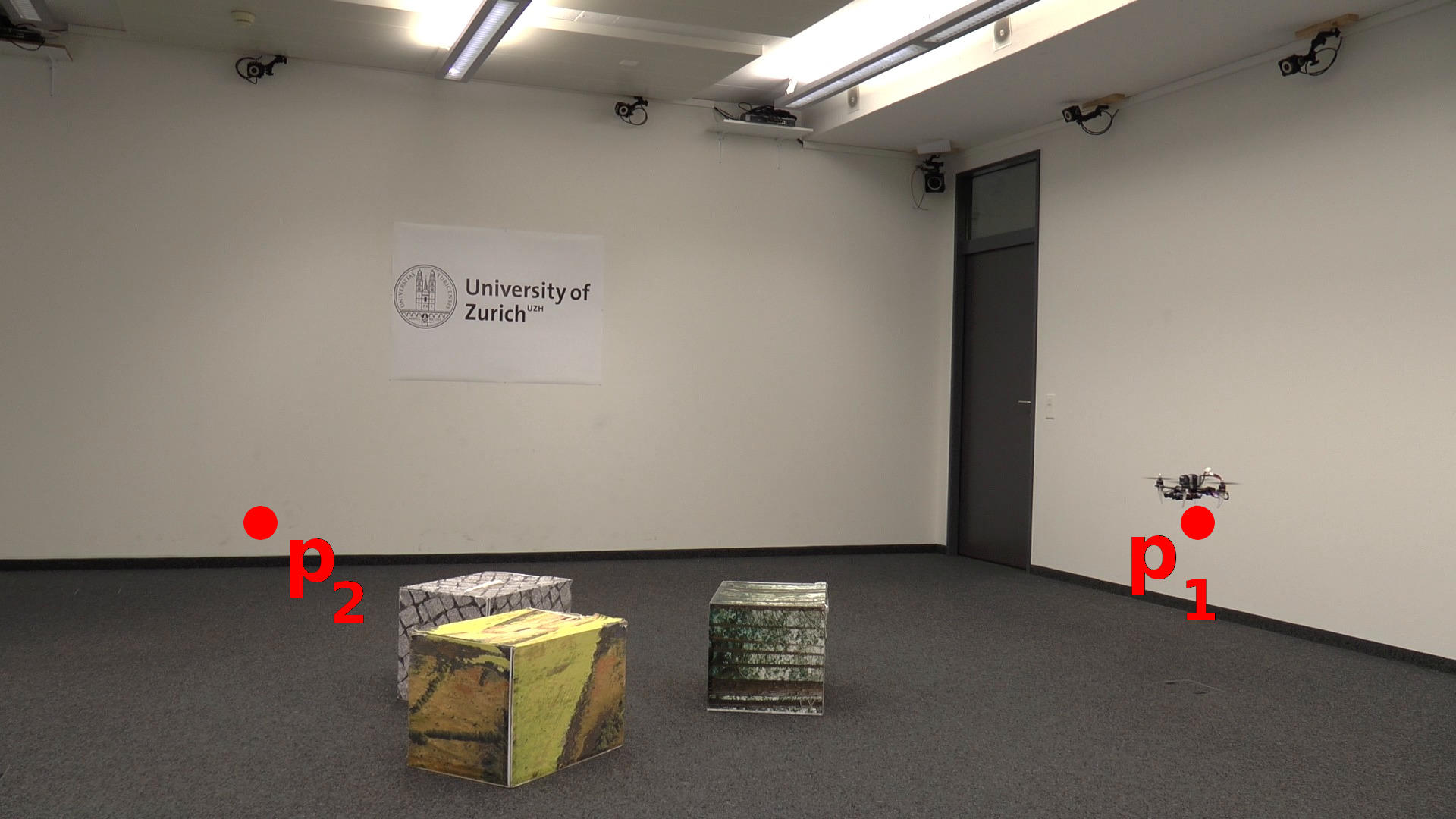}\hfill
	\includegraphics[width=.21\textwidth,trim={5cm 1cm 7cm 7cm},clip]{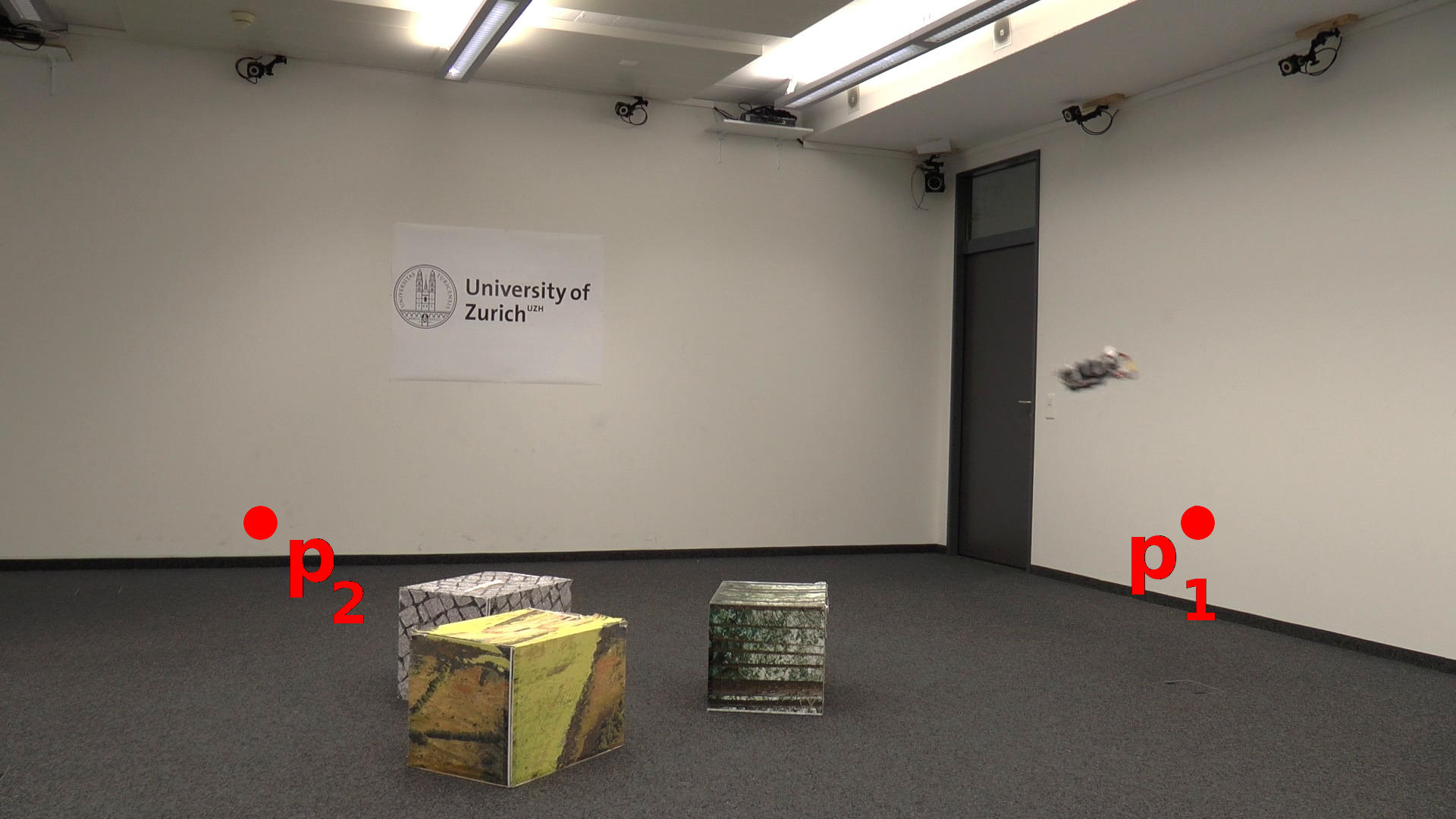}\hfill
	\includegraphics[width=.21\textwidth,trim={5cm 1cm 7cm 7cm},clip]{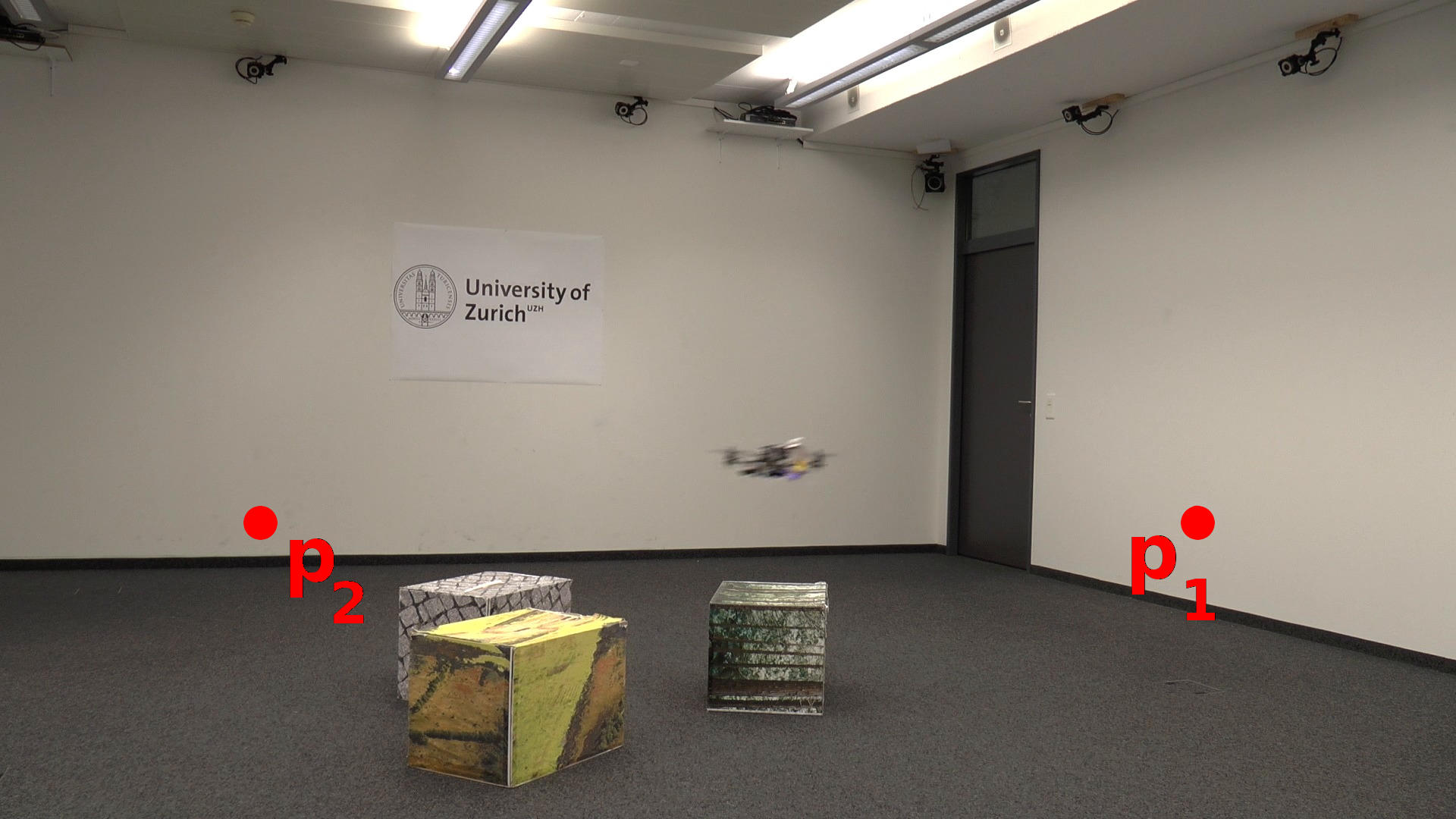}\hfill
	\includegraphics[width=.21\textwidth,trim={5cm 1cm 7cm 7cm},clip]{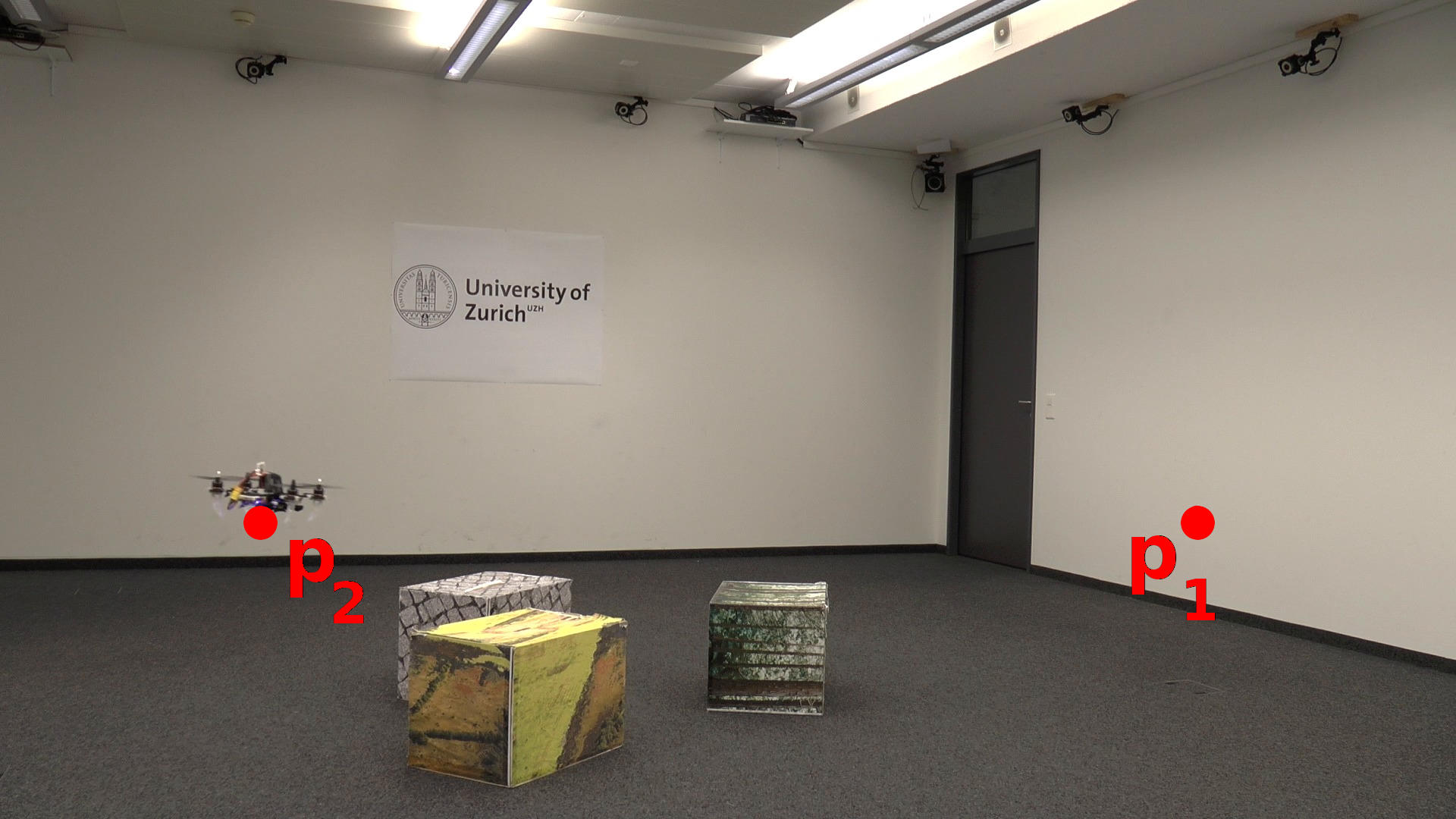}
	
	\medskip
	
	\includegraphics[width=.21\textwidth]{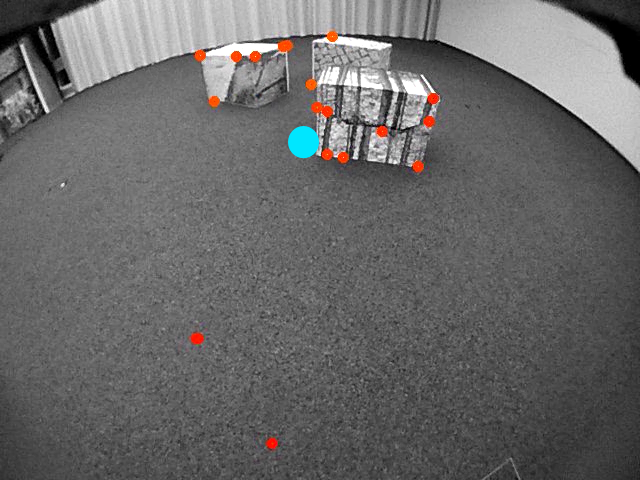}\hfill
	\includegraphics[width=.21\textwidth]{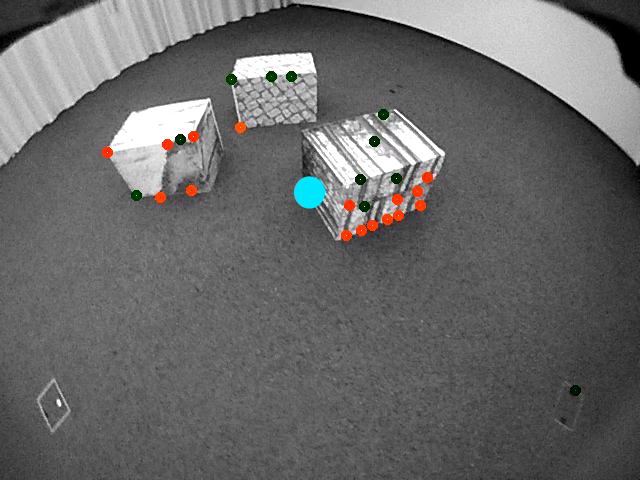}\hfill
	\includegraphics[width=.21\textwidth]{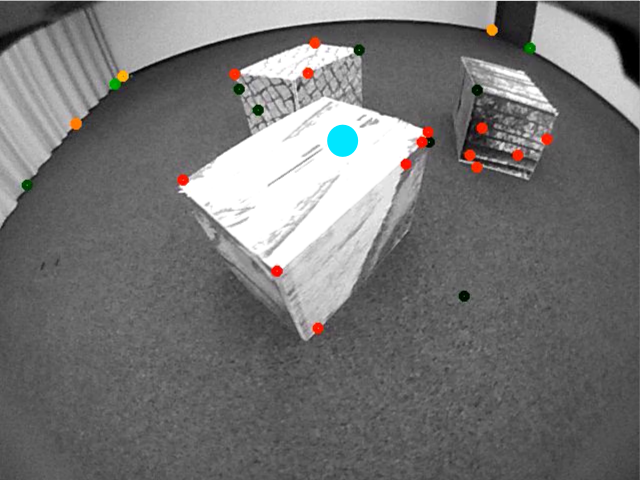}\hfill
	\includegraphics[width=.21\textwidth]{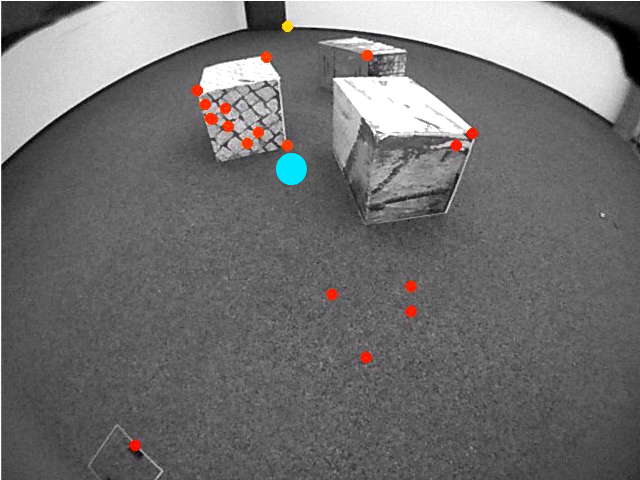}
	\caption{A sequence of the visibility experiment for the hover-to-hover flight experiment, with time progressing from left to right. The quadrotor performs a maneuver to fly to a new reference pose, exploiting additional height to pitch less and keep the point of interest (centroid of the vision features, marked as cyan circle) in the center of the image. The corresponding footage is available in the accompanying video.}
	\label{fig:box_sequence}
\end{figure*}
\begin{figure*}[t]
	\centering
	\subfloat[\label{fig:reprojection_circle}Reprojection for the circular trajectory.]{\includegraphics[width=.33\textwidth,trim={0 0 0 12},clip]{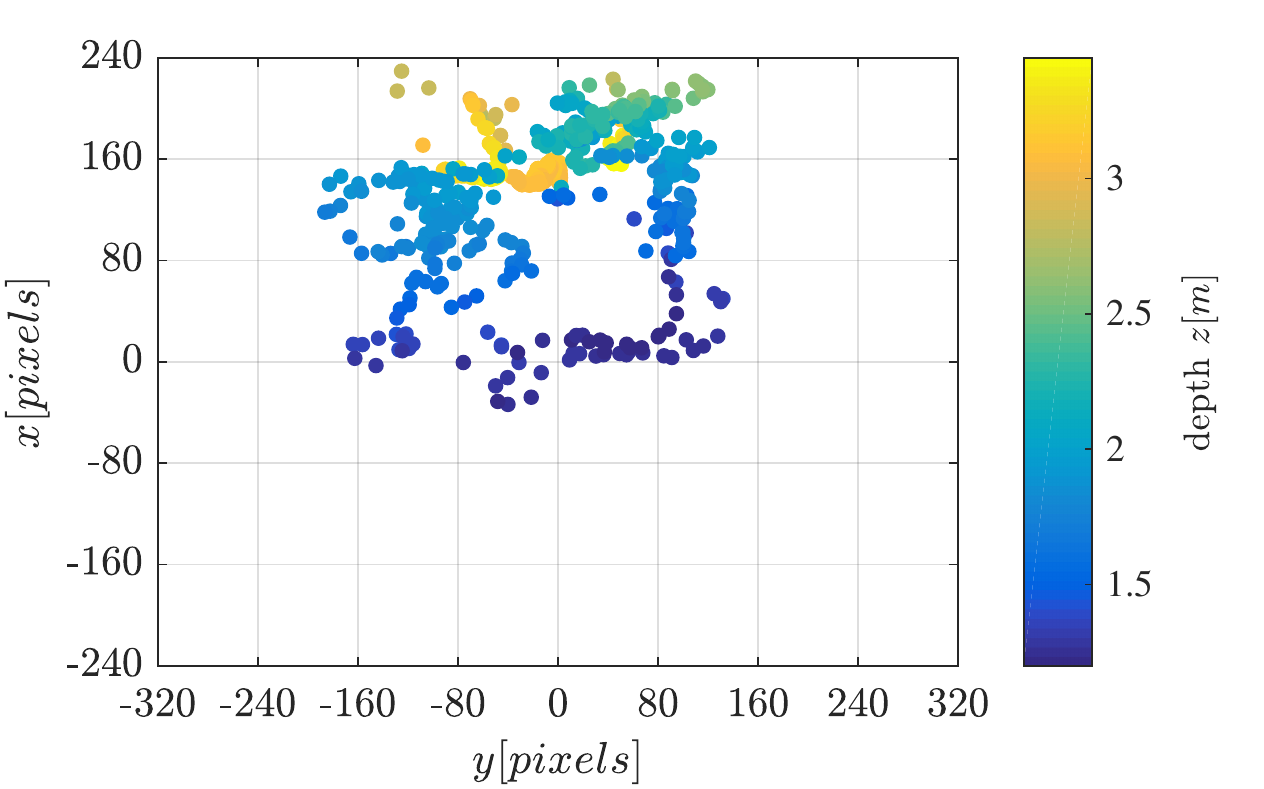}} \hfill
	\subfloat[\label{fig:reprojection_boxes}Reprojection for the \emph{hover-to-hover} experiment.]{\includegraphics[width=.33\textwidth,trim={0 0 0 12},clip]{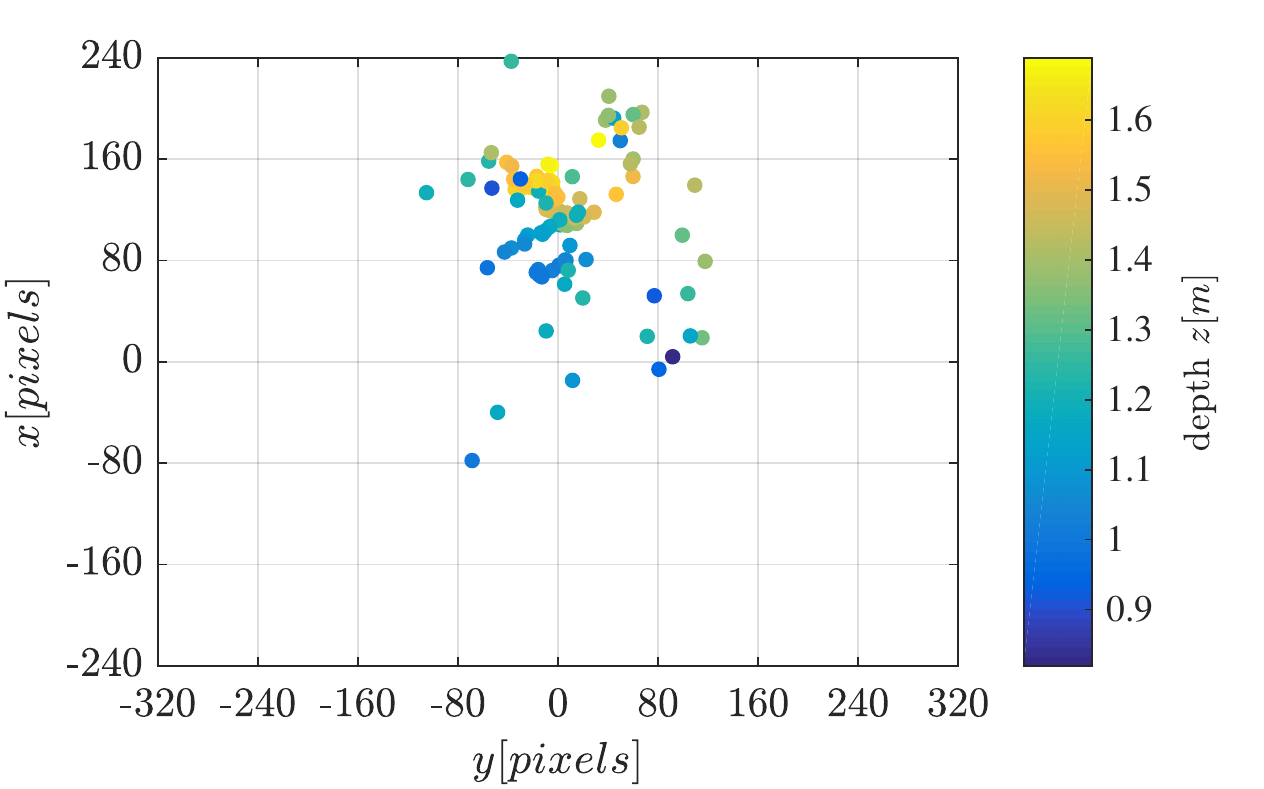}} \hfill
	\subfloat[\label{fig:reprojection_darkness}Reprojection for the darkness experiment.]{\includegraphics[width=.33\textwidth,trim={0 0 0 12},clip]{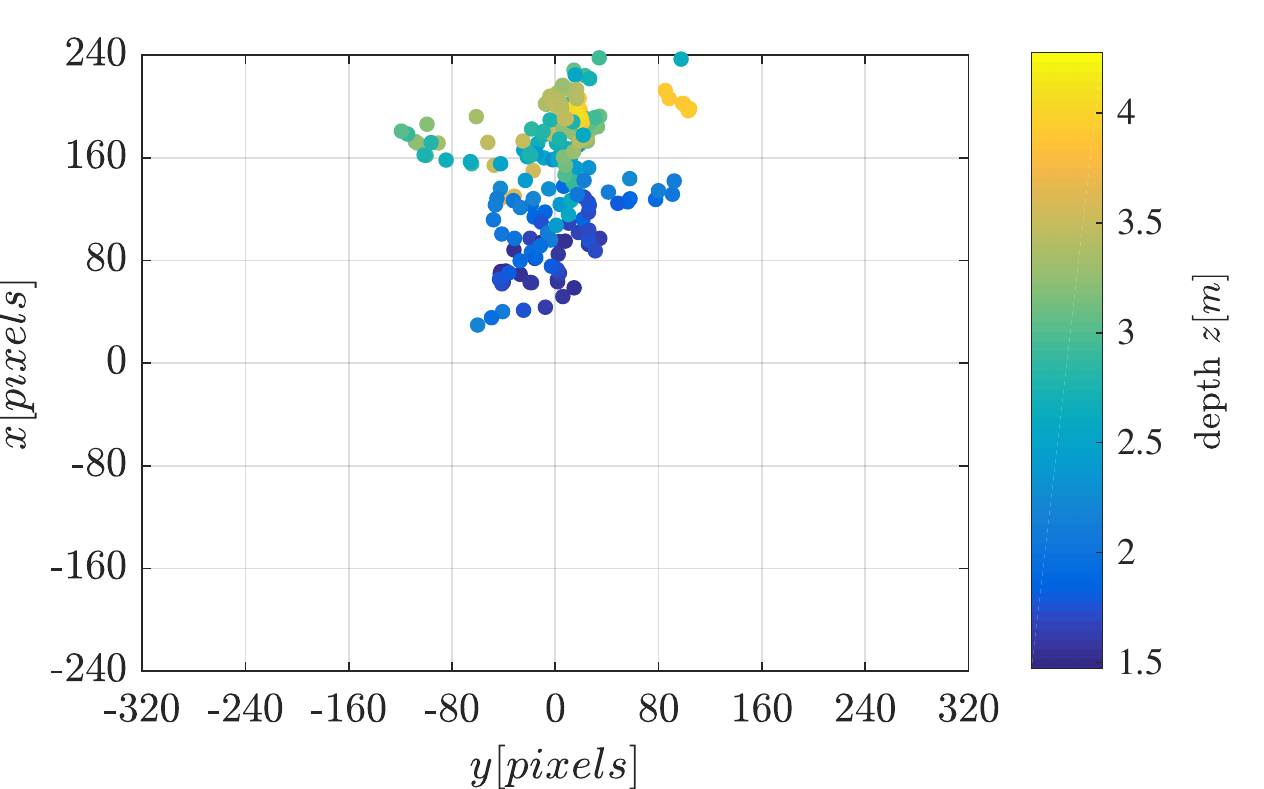}}
	\caption{Reprojection of the point of interest in image plane, colored according to the depth with respect to the camera frame.}
	\label{fig:reprojection}
\end{figure*}

\subsection{Experimental Setup} \label{sec:experimental_setup}
We used a small and lightweight quadrotor platform to achieve high agility through high torque-to-inertia and thrust-to-weight ratios, and improve simplicity and safety for the user (cf. Fig \ref{fig:quadrotor_closeup}).
The quadrotor had a take-off weight of $\SI{420}{\gram}$, a thrust-to-weight ratio of $\sim 2$, and a motor-to-motor diagonal of $\SI{220}{\milli \meter}$.
We used a Qualcomm Snapdragon Flight board with a quad-core ARM processor at up to $\SI{2.26}{\giga\hertz}$ and $\SI{2}{\giga\byte}$ of RAM, paired with a Qualcomm Snapdragon Flight ESC.
The board was equipped with an Inertial Measurement Unit and a forward-looking, wide field-of-view global-shutter camera tilted down by \SI{45}{\degree} for visual-inertial odometry (VIO) using the Qualcomm mvSDK.
It ran ROS on Linux and our self-developed flight stack.
We setup the optimization with ACADO and used qpOASES as solver.
As discretization step, we chose $dt = \SI{0.1}{\second}$ with a time horizon of $t_h = \SI{2}{\second}$ and ran one iteration step in each control loop with a frequency of $\SI{100}{\hertz}$.
Therefore, the iteration ran roughly $10\times$ faster than the discretization time, resulting in small deviations of the predicted state vector between iterations and facilitating convergence.
The code developed in this work is publicly available as open-source software.

\subsection{Experiment Description and Results}
To prove the functionality and importance of our PAMPC, we ran three experiments.
In the first experiment, the controller modified a circular trajectory to improve the visibility of a point of interest.
In the second experiment, the controller handled \emph{hover-to-hover} flight by deviating from a straight line trajectory to keep the point of interest visible.
In the third experiment, it enabled vision-based flight in an extremely challenging scenario.
All the experiments were conducted with onboard VIO and onboard computation of the PAMPC, without any offline computation and without any motion-capture system.

\subsubsection{Circular Flight}
\label{sec:experiment_circle}
We setup a small pile of boxes in the middle of a room otherwise poor of texture.
We did this to force the VIO pipeline to use such boxes as features for state estimation.
The centroid of these features was set as our point of interest.
We provided the robot with a circular reference trajectory around the aforementioned boxes and asked it to fly along such a trajectory while maintaining the boxes visible in the center of the image (cf. Fig.~\ref{fig:eye_catching}).
We evaluated the performance of our framework for speeds along the circle from \SI{1}{\meter \per \second} to \SI{3}{\meter \per \second}.

\begin{figure}[t]
	\centering
	\includegraphics[width=0.9\linewidth, trim={11 17 25 35},clip]{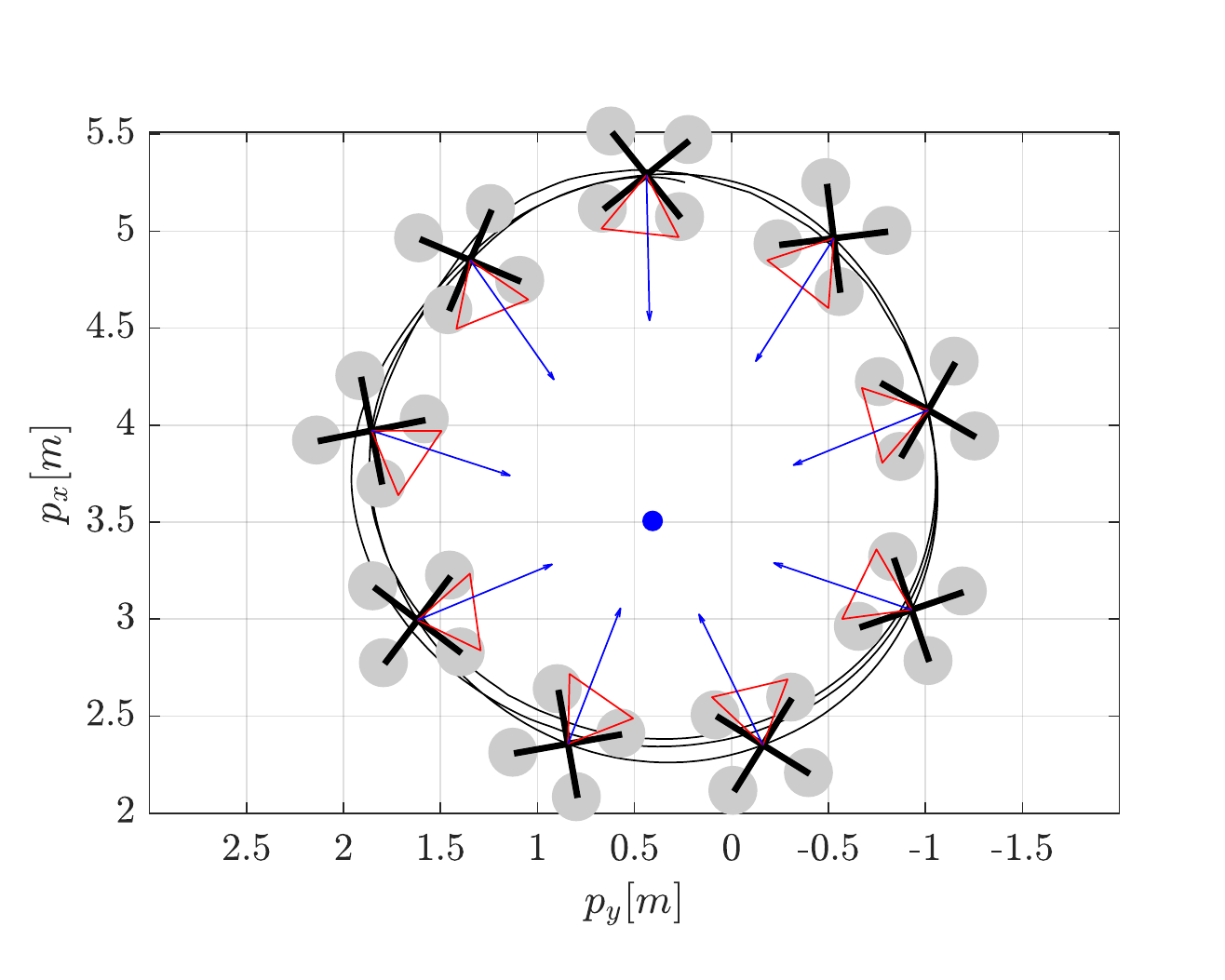}
	\caption{Executed trajectory with quadrotor heading while the arrow points toward the point of interest (blue).}
	\label{fig:circle_flight_results}
\end{figure}

The results of one run of the circular flight experiments at \SI{3}{\meter \per \second} are depicted in Fig.~\ref{fig:circle_flight_results}.
Despite the agility of the maneuver, which requires large deviations from the hover conditions, the robot is able to keep the point of interest visible in the onboard image.
Fig.~\ref{fig:reprojection_circle} reports the reprojection in the image plane for such point of interest.

\subsubsection{Hover-To-Hover Flight}
\label{sec:experiment_visibility}
In this experiment within the same scenario as in Sec.~\ref{sec:experiment_circle}, we showed the capabilities of our framework for \emph{hover-to-hover} flight.
More specifically, we requested a pose jump from a position $\bVec{p}_1$ to $\bVec{p}_2$ at equal height (cf. Fig.~\ref{fig:box_sequence}).
During that maneuver, the quadrotor had to pitch down to reach the desired acceleration, but controversially should pitch as little as possible to keep the point of interest visible.
A sequence of this experiment is visible in Fig.~\ref{fig:box_sequence}.

\begin{figure}[t]
	\centering
	\includegraphics[width=0.9\linewidth, trim={0 30 25 50},clip]{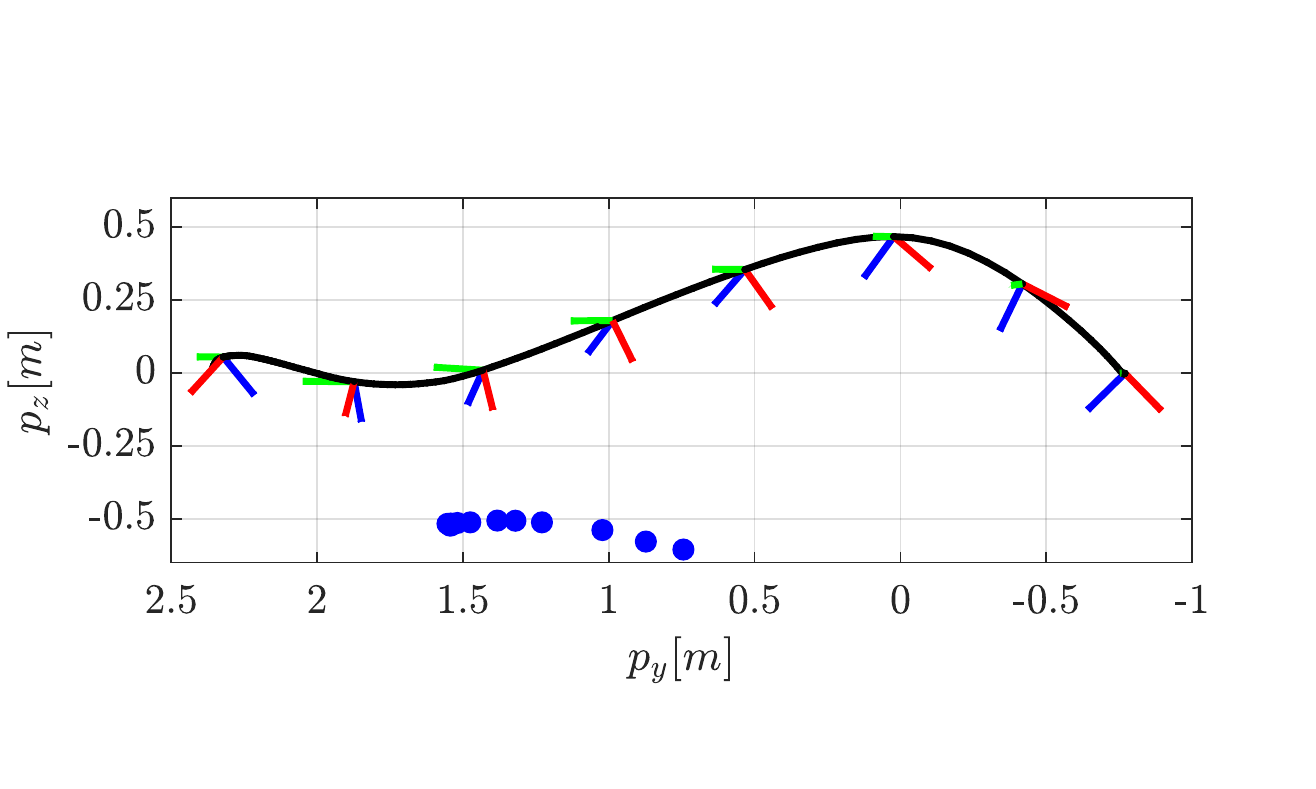}
	\medskip
	\includegraphics[width=0.9\linewidth, trim={0 45 25 70},clip]{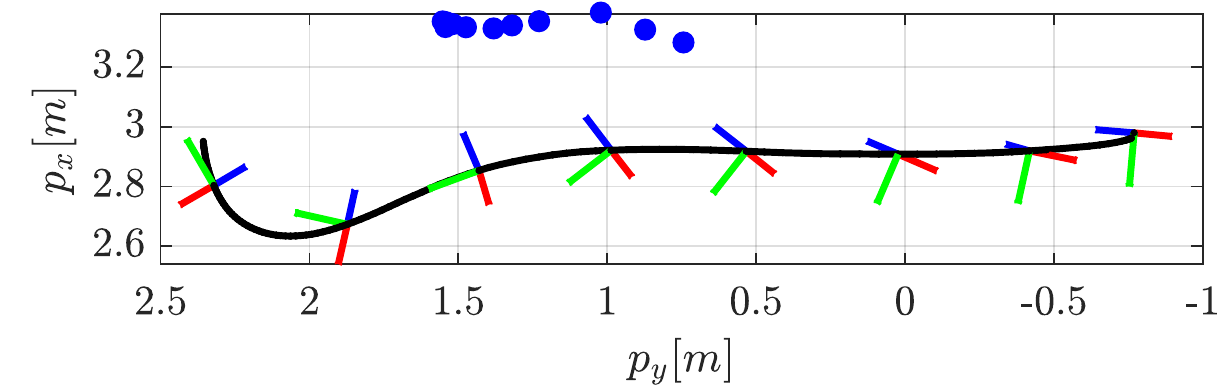}
	\caption{Quadrotor path in hover-to-hover, looking towards the centroid of tracked features (blue), with the camera frame indicated by $\{${\color{red}$x_C$}, {\color{green}$y_C$}, {\color{blue}$z_C$}$\}$.}
	\label{fig:path_boxes}
\end{figure}
One can easily see that, despite the start and end positions are at the same height, the quadrotor not only pitches to go towards the new reference in an horizontal motion, but also accelerates upward (i.e., in positive $z$, cf. Fig.~\ref{fig:path_boxes}).
This results in a smaller pitch angle and a higher thrust to reach the same $y$-acceleration, which is helpful for perception since it brings the features towards the center of the frame due to the higher altitude.
If perception objectives were not considered, the resulting trajectory would have not required any height change, potentially leading to a poor visibility of the point of interest.
The full motion of the quadrotor is depicted in Fig.~\ref{fig:path_boxes}, where the exploitation of the added height and the orientation of the camera frame can be seen.
Finally, in Fig.~\ref{fig:reprojection_boxes} we show the reprojection in the image plane for the point of interest.

\subsubsection{Darkness Scenario}
\label{sec:experiment_environment}
\begin{figure}[t!]
	\centering
	\includegraphics[width=.23\textwidth,trim={10cm 10cm 18cm 12cm},clip]{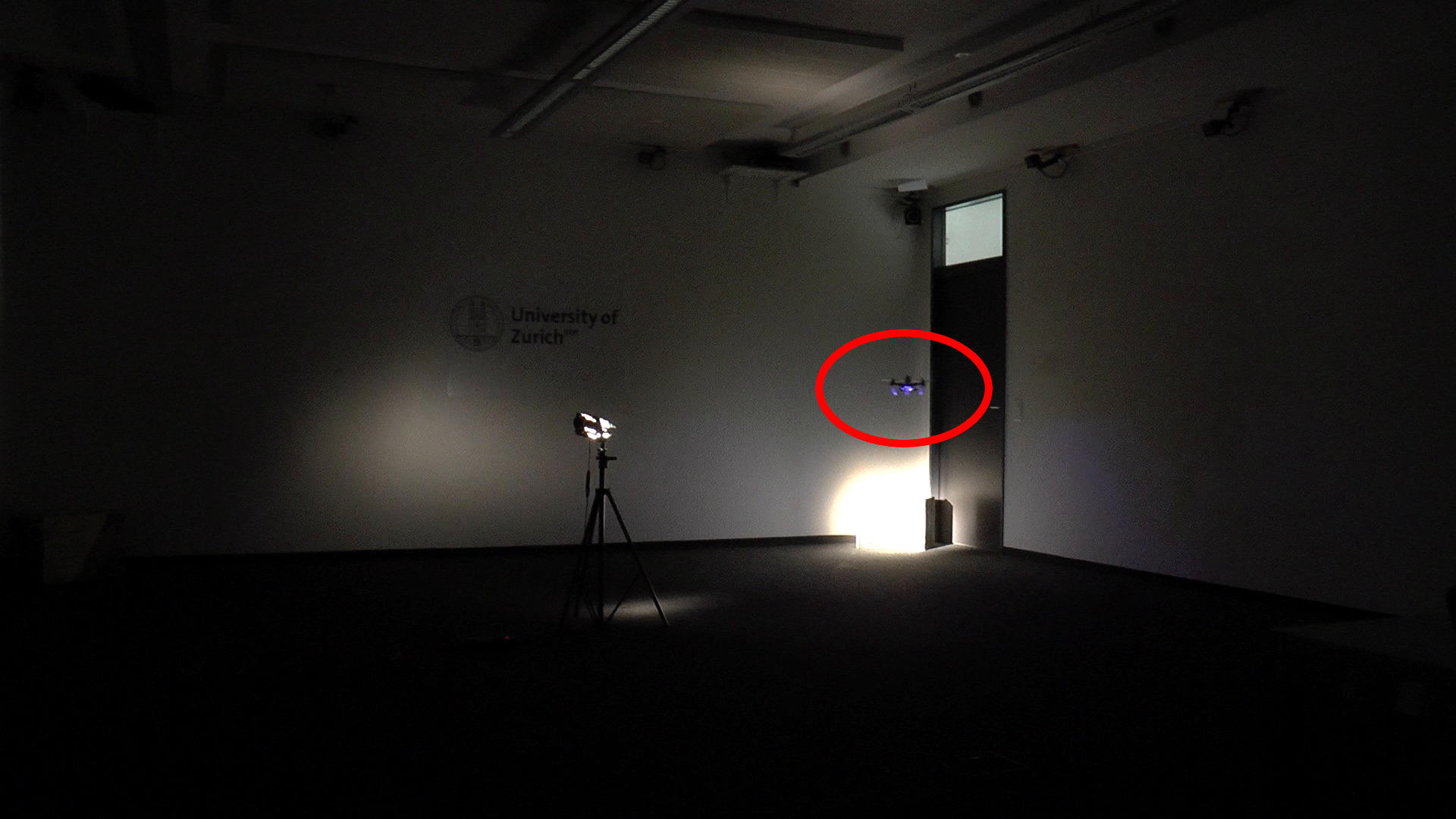}\hfill
	\includegraphics[width=.23\textwidth,trim={10cm 10cm 18cm 12cm},clip]{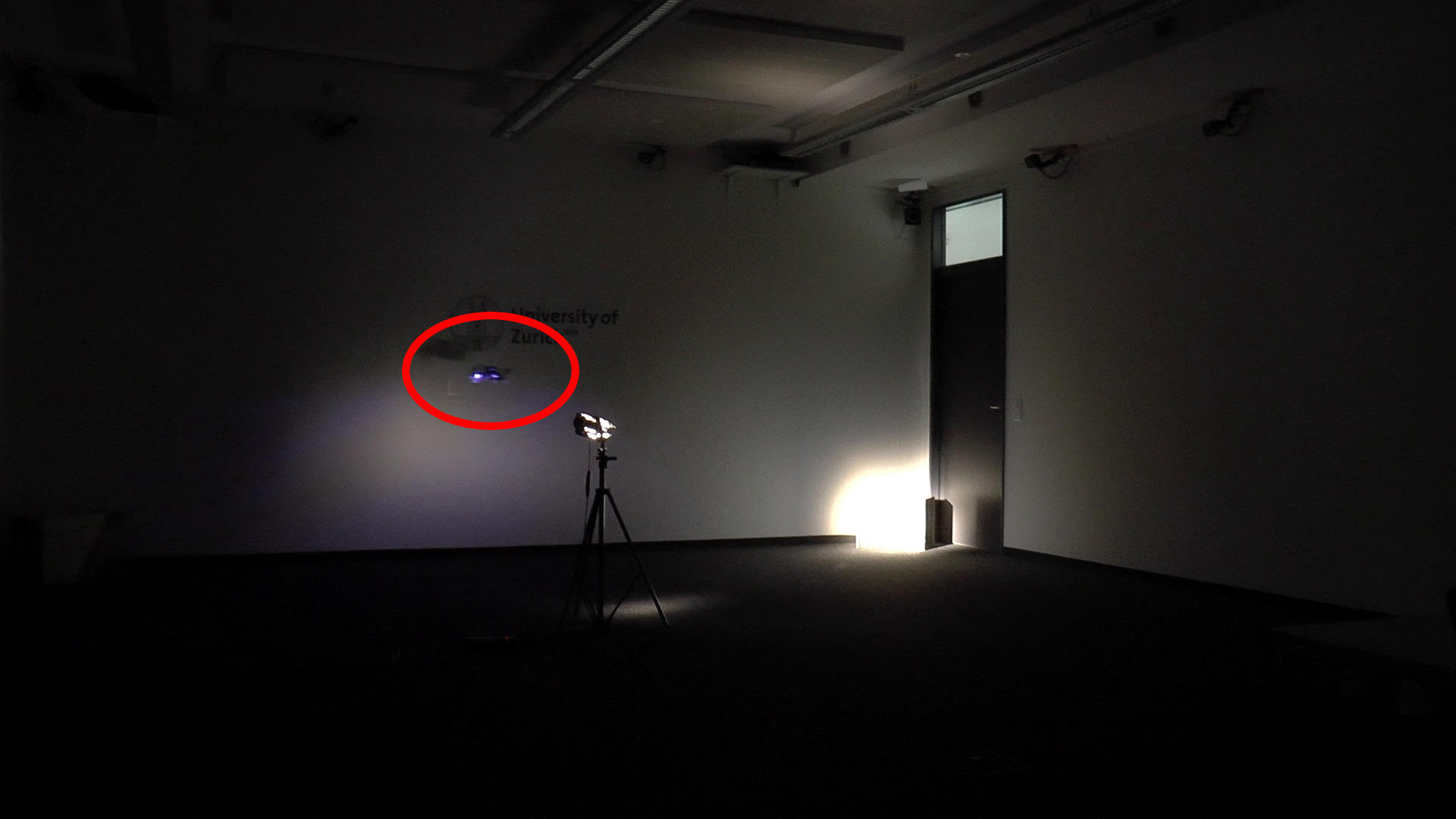}\hfill
	
	\medskip
	
	\includegraphics[width=.23\textwidth,trim={0 120 0 0},clip]{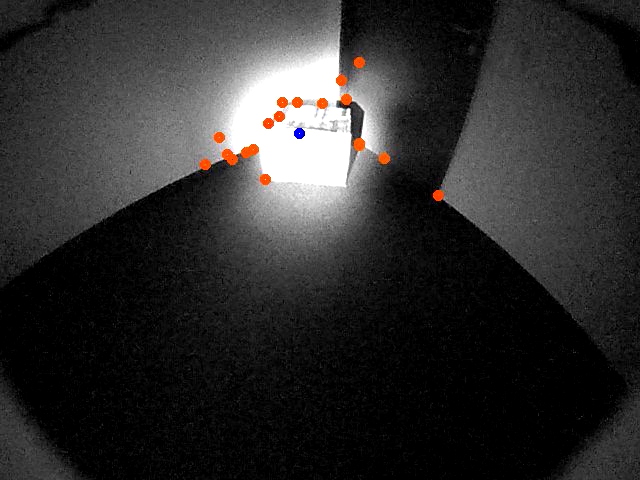}\hfill
	\includegraphics[width=.23\textwidth,trim={0 120 0 0},clip]{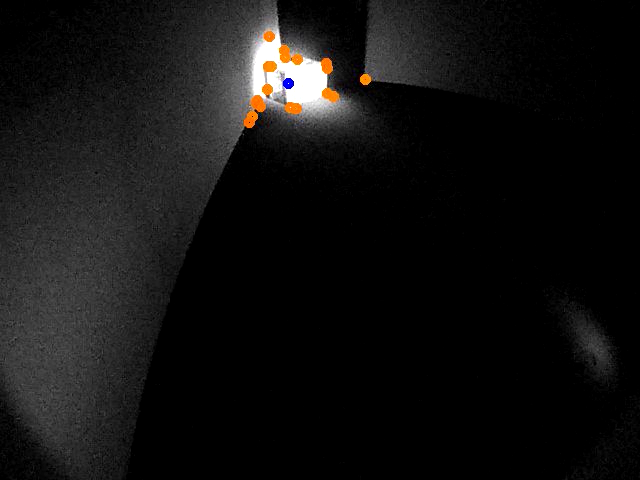}\hfill
	\caption{A sequence of the darkness experiment with time progressing from left to right. The quadrotor, highlighted by a red circle in the figures in the first row, tracks a trajectory and adjusts its heading to keep the point of interest (centroid of the vision features) in field of view.}
	\label{fig:spotlight_sequence}
\end{figure}

This experiment was targeted towards extremely challenging scenarios, such as flight in a very dark environment, or otherwise difficult illumination conditions (cf. Fig.~\ref{fig:spotlight_sequence}).
To demonstrate the performance in such a scenario, we flew the vehicle several times in a dark room with two illuminated spots.
If the illuminated spots left the field of view for a moment, the VIO pipeline would drift quickly or even completely loose track, potentially leading to a crash.
Therefore, in such scenarios it was of immense importance to keep the few available features always visible.
The flown path was given by a trajectory passing through four waypoints forming a rectangle, but without any heading reference.
\begin{figure}[t]
	\centering
	\includegraphics[width=0.8\linewidth, trim={20 0 45 20},clip]{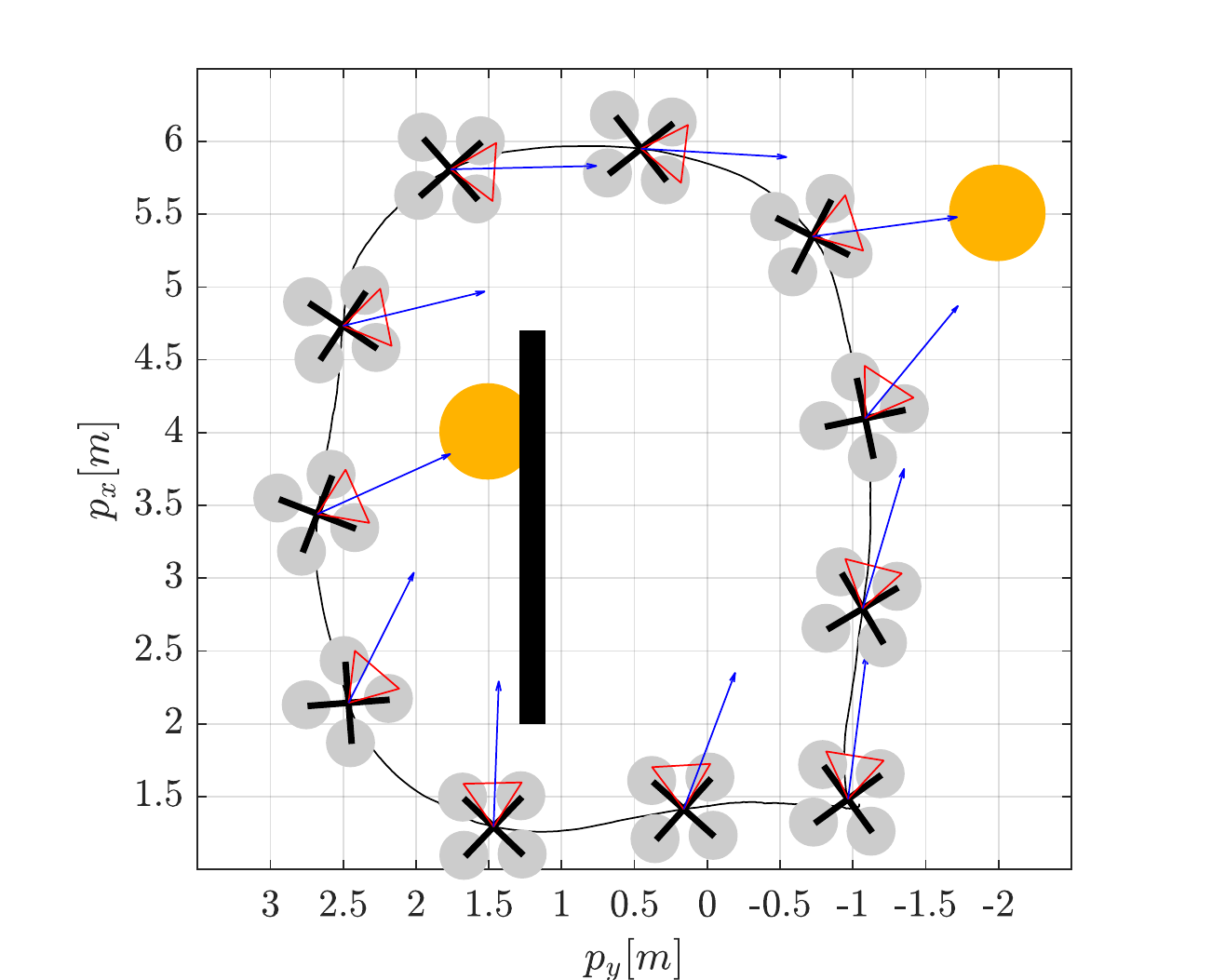}
	\caption{Path of the quadrotor, looking towards light spots (yellow), with camera direction (red triangle) and point-of-interest direction (blue arrow).}
	\label{fig:path_spotlights}
\end{figure}
The quadrotor correctly adjusts its heading to keep the illuminated spot in its field of view, because this is the only source of trackable features.
Fig.~\ref{fig:path_spotlights} visualizes a setup with two spotlights and a cardboard wall in between, where the quadrotor first focuses on the upper right illuminated spot, and further down the track switches to the second illuminated spot behind the wall.
The reprojection of the point of interest in the image plane is shown in Fig.~\ref{fig:reprojection_darkness}.

%% file: chapters/discussion.tex
\section{Discussion} \label{sec:discussion}

\subsection{Choice of the optimizer}
To implement the optimization problem, we chose to use ACADO because of two main reasons:
(\rom{1}) it is capable of transcribing system dynamics with single- and multiple-shooting and integration schemes, as well as provide an interface to a solver;
(\rom{2}) it generates c++ code, which then is compiled directly on the executing platform, which allows it to use accelerators and optimizations tailored to the platform.

\subsection{Convexity of the problem}
Our state and input space is a convex domain, hence also any quadratic cost in those is convex.
The perception costs could be argued to be non-convex due to the division by $\featurePosInCameraz$ in the projection \eqref{eq:camera_projection_model}.
However, on examination of the projection one will notice that the denominator $\featurePosInCameraz$ is always positive, since the pinhole camera projection model does not allow negative or zero depths.
We can therefore constrain $\featurePosInCameraz$ to be positive, rendering all possible solutions in the positive halfplane $\mathcal{R}^+$ and therefore recover convexity.

\subsection{Choice of point of interest}
In our experiments, we used the centroid of detected features as our point of interest.
Assuming that all the features are equally important, instead of optimizing for each individually, we can summarize them as their centroid, which results in the same optimal solution.

\subsection{PAMPC Parameters}
We chose a discretization of ${dt = \SI{0.1}{\second}}$ and a time horizon of ${t_h = \SI{2}{\second}}$.
One could always argue that a longer time horizon and a shorter discretization step are beneficial, but they also increase the computation time by roughly $\mathcal{O}(N^2)$ with the number of discretization nodes ${N= \frac{t_h}{dt}}$.
In our experience, we could not identify any significant gain from smaller discretization steps nor from a longer time horizon.

\subsection{Computation Time}
Since the computation time must be low enough to execute the optimization in a real time scheme, we show that it is significantly lower than the one required by the controller frequency of $\SI{100}{\hertz}$.
Indeed, our PAMPC requires on average $\SI{3.53}{\milli\second}$.
It is interesting to note that this is the case for both an idle CPU and while running the full pipeline with VIO and our full control pipeline.
This is due to the quad-core ARM CPU and the fact that our full pipeline without the PAMPC takes up only 3 cores leaving one free for the PAMPC.
However, the standard deviation increases significantly if the CPU is under load (from \SI{0.155}{\milli\second} to \SI{0.354}{\milli\second}), even though the maximal execution time always stays below $\SI{5}{\milli\second}$.

\subsection{Drawbacks of a Two-Step Approach}
An alternative approach to the problem tackled in this work is to use the differential flatness as in \cite{Mellinger11icra} to plan a translational trajectory connecting the start and end positions, and subsequentially plan the yaw angle to point the camera towards the point of interest.
After planning, a suitable controller could be used to track the desired reference trajectory.
Although possible, such a solution would lead to sub-optimal results because of the following reasons:
(\rom{1}) the roll and pitch angles of the quadrotor would be planned without considering the visibility objective, therefore might render the point of interest not visible in the image despite the yaw control;
(\rom{2}) because of the split between planning and control, even if the first would provide guarantees about visibility, these could not be preserved during the control stage due to deviations from the nominal trajectory;
(\rom{3}) it would be challenging to provide guarantees about the respect of the input saturations.
Therefore, our proposed approach considering perception, planning and control as a single problem leads to superior results.

%% file: chapters/conclusions.tex
\section{Conclusions} \label{sec:conclusions}
In this work, we presented a perception-aware model predictive control (PAMPC) algorithm for quadrotors able to optimize both action and perception objectives.
Our framework computes trajectories that satisfy the system dynamics and inputs limits of the platform. 
Additionally, it optimizes perception objectives by maximizing the visibility of a point of interest in the image and minimizing the velocity of its projection into the image plane for robust and reliable sensing.
To fully exploit the agility of a quadrotor, we incorporated perception objectives into the optimization problem not as constraints, but rather as components in the cost function to be optimized.
Our algorithm is able to run in real-time on an onboard ARM processor, in parallel with a VIO pipeline, and is used to directly control the robot.
We validated our approach in real-world experiments using a small-scale, lightweight quadrotor platform.

%% file: iros2018_perception_aware_mpc.bbl
\begin{thebibliography}{10}
\providecommand{\url}[1]{#1}
\csname url@samestyle\endcsname
\providecommand{\newblock}{\relax}
\providecommand{\bibinfo}[2]{#2}
\providecommand{\BIBentrySTDinterwordspacing}{\spaceskip=0pt\relax}
\providecommand{\BIBentryALTinterwordstretchfactor}{4}
\providecommand{\BIBentryALTinterwordspacing}{\spaceskip=\fontdimen2\font plus
\BIBentryALTinterwordstretchfactor\fontdimen3\font minus
  \fontdimen4\font\relax}
\providecommand{\BIBforeignlanguage}[2]{{%
\expandafter\ifx\csname l@#1\endcsname\relax
\typeout{** WARNING: IEEEtran.bst: No hyphenation pattern has been}%
\typeout{** loaded for the language `#1'. Using the pattern for}%
\typeout{** the default language instead.}%
\else
\language=\csname l@#1\endcsname
\fi
#2}}
\providecommand{\BIBdecl}{\relax}
\BIBdecl

\bibitem{Mellinger10iser}
D.~Mellinger, N.~Michael, and V.~Kumar, ``Trajectory generation and control for
  precise aggressive maneuvers with quadrotors,'' in \emph{Int. Symp.
  Experimental Robotics}, 2010.

\bibitem{Mellinger11icra}
D.~Mellinger and V.~Kumar, ``Minimum snap trajectory generation and control for
  quadrotors,'' in \emph{{IEEE} Int. Conf. Robot. Autom.}, 2011.

\bibitem{Mueller15tro}
M.~W. Mueller, M.~Hehn, and R.~D'Andrea, ``A computationally efficient motion
  primitive for quadrocopter trajectory generation,'' \emph{{IEEE} Trans.
  Robot.}, vol.~31, no.~6, 2015.

\bibitem{Brescianini13iros}
D.~Brescianini, M.~Hehn, and R.~D'Andrea, ``Quadrocopter pole acrobatics,'' in
  \emph{IEEE/RSJ Int. Conf. Intell. Robot. Syst.}, 2013.

\bibitem{Falanga17icra}
D.~Falanga, E.~Mueggler, M.~Faessler, and D.~Scaramuzza, ``Aggressive quadrotor
  flight through narrow gaps with onboard sensing and computing,'' in
  \emph{{IEEE} Int. Conf. Robot. Autom.}, 2017.

\bibitem{Loianno17ral}
G.~Loianno, C.~Brunner, G.~McGrath, and V.~Kumar, ``Estimation, control, and
  planning for aggressive flight with a small quadrotor with a single camera
  and {IMU},'' \emph{{IEEE} Robot. Autom. Lett.}, vol.~2, no.~2, 2017.

\bibitem{Su17iser}
K.~Su and S.~Shen, ``Catching a flying ball with a vision-based quadrotor,'' in
  \emph{Int. Symp. Experimental Robotics}, 2016.

\bibitem{Kamel16Arxiv}
\BIBentryALTinterwordspacing
M.~Kamel, M.~Burri, and R.~Siegwart, ``Linear vs nonlinear {MPC} for trajectory
  tracking applied to rotary wing micro aerial vehicles,'' \emph{arXiv}, 2016.
  [Online]. Available: \url{http://arxiv.org/abs/1611.09240}
\BIBentrySTDinterwordspacing

\bibitem{Neunert16icra}
M.~Neunert, C.~de~Crousaz, F.~Furrer, M.~Kamel, F.~Farshidian, R.~Siegwart, and
  J.~Buchli, ``Fast nonlinear model predictive control for unified trajectory
  optimization and tracking,'' in \emph{{IEEE} Int. Conf. Robot. Autom.}, 2016.

\bibitem{Bangura14ifac}
M.~Bangura and R.~Mahony, ``Real-time model predictive control for
  quadrotors,'' \emph{{IFAC} World Congress}, 2014.

\bibitem{Penin17iros}
B.~Penin, R.~Spica, P.~Robuffo~Giordano, and F.~Chaumette, ``{Vision-Based
  Minimum-Time Trajectory Generation for a Quadrotor UAV},'' in \emph{IEEE/RSJ
  Int. Conf. Intell. Robot. Syst.}, 2017.

\bibitem{Spica17ijrr}
R.~Spica, P.~Robuffo~Giordano, and F.~Chaumette, ``{Coupling Active Depth
  Estimation and Visual Servoing via a Large Projection Operator},'' \emph{Int.
  J. Robot. Research}, vol.~36, no.~11, 2017.

\bibitem{Costante17ISRR}
G.~Costante, J.~Delmerico, M.~Werlberger, P.~Valigi, and D.~Scaramuzza,
  ``Exploiting photometric information for planning under uncertainty,''
  \emph{Robotics Research: Volume 1}, pp. 107--124, 2018.

\bibitem{Forster14rss}
C.~Forster, M.~Pizzoli, and D.~Scaramuzza, ``Appearance-based active,
  monocular, dense depth estimation for micro aerial vehicles,'' in
  \emph{Robotics: Science and Systems}, 2014.

\bibitem{Sheckells16iros}
M.~Sheckells, G.~Garimella, and M.~Kobilarov, ``Optimal visual servoing for
  differentially flat underactuated systems,'' in \emph{IEEE/RSJ Int. Conf.
  Intell. Robot. Syst.}, 2016.

\bibitem{Naegeli17ral}
T.~N{\"a}geli, J.~Alonso-Mora, A.~Domahidi, D.~Rus, and O.~Hilliges,
  ``Real-time motion planning for aerial videography with dynamic obstacle
  avoidance and viewpoint optimization,'' \emph{{IEEE} Robot. Autom. Lett.},
  vol.~2, no.~3, 2017.

\bibitem{Naegeli17siggraph}
T.~N{\"a}geli, L.~Meier, A.~Domahidi, J.~Alonso-Mora, and O.~Hilliges,
  ``Real-time planning for automated multi-view drone cinematography,'' in
  \emph{SIGGRAPH}, 2017.

\bibitem{Potena17ecmr}
C.~Potena, D.~Nardi, and A.~Pretto, ``Effective target aware visual navigation
  for uavs,'' in \emph{Eur. Conf. Mobile Robots}, 2017.

\bibitem{Szeliski10book}
R.~Szeliski, \emph{Computer Vision: Algorithms and Applications}, 2010.

\bibitem{Houska2011b}
B.~{H}ouska, H.~Ferreau, and M.~Diehl, ``{An Auto-Generated Real-Time Iteration
  Algorithm for Nonlinear {MPC} in the Microsecond Range},'' \emph{Automatica},
  vol.~47, no.~10, 2011.

\bibitem{Houska2011a}
B.~Houska, H.~Ferreau, and M.~Diehl, ``{ACADO} {T}oolkit -- {A}n {O}pen
  {S}ource {F}ramework for {A}utomatic {C}ontrol and {D}ynamic
  {O}ptimization,'' \emph{Optimal Control Applications and Methods}, vol.~32,
  no.~3, 2011.

\end{thebibliography}
